\documentclass[lettersize,journal]{IEEEtran}
\usepackage{amsmath,amsfonts}
\usepackage{algorithmic}
\usepackage{algorithm}
\usepackage{array}
\usepackage[justification=centering]{caption}
\usepackage{subfigure}
\usepackage{textcomp}
\usepackage{stfloats}
\usepackage{url}
\usepackage{multirow}
\usepackage{verbatim}
\usepackage{graphicx}
\usepackage{cite}
\usepackage{amssymb}          
\usepackage{mathrsfs}
\usepackage{xcolor}           
\usepackage{algorithm}

\usepackage{bm}
\begin{document}
\title{Robust Unsupervised Fault Diagnosis For High-Dimensional Nonlinear Noisy Data}
\author{Dandan Zhao, Hongpeng Yin, Jintang Bian, Han Zhou
\thanks{Dandan Zhao and Hongpeng Yin are with the School of Automation, Chongqing University, Chongqing, 400044, China (e-mail: whsmhgy@gmail.com, yinhongpeng@gmail.com).

Jingtang Bian is with the School of Computer Science, Sun Yat-Sen University,  Guangzhou, 510006, China.  (E-mail: bianjt@mail2.sysu.edu.cn).

Han Zhou is with the Department of New Networks, Peng Cheng Laboratory, Shenzhen, China.
(zhouhan1515@foxmail.com )


This work was supported in part by the National Natural Science Foundation of China under Grant 62273062, in part by the Natural Science Foundation of Chongqing under Grant 2022NSCQ-LZX0324, in part by the Chongqing Talents: Exceptional Young Talents Project under Grant cstc2021ycjh-bgzxm0028, and by the Major Key Project of PCL (Grant No. PCL2024A05)}}

\markboth{IEEE Internet of Things Journal}%
{Shell \MakeLowercase{\textit{Zhao et al.}}: Adaptive Similarity-Aware for Robust Unsupervised Discriminative Projection}


\maketitle

\begin{abstract}
Traditional fault diagnosis methods struggle to handle fault data, with complex data characteristics such as high dimensions and large noise. Deep learning is a promising solution, which typically works well only when labeled fault data are available. To address these problems, a robust unsupervised fault diagnosis using machine learning is proposed in this paper. First, a special dimension reduction method for the high-dimensional fault data is designed. Second, the extracted features are enhanced by incorporating nonlinear information through the learning of a graph structure. Third, to alleviate the problem of reduced fault-diagnosis accuracy attributed to noise and outliers, $l_{2,1}$-norm and typicality-aware constraints are introduced from the perspective of model optimization, respectively. Finally, this paper provides comprehensive theoretical and experimental evidence supporting the effectiveness and robustness of the proposed method. The experiments on both the benchmark Tennessee-Eastman process and a real hot-steel milling process show that the proposed method exhibits better robustness compared to other methods, maintaining high diagnostic accuracy even in the presence of outliers or noise.
\end{abstract}



\section{Introduction}
Fault diagnosis, detecting the presence of faults, and identifying the types of faults in an industrial system, play an irreplaceable role in ensuring production safety and reducing the profitability of modern industry processes \cite{ref1, ref2, ref3}. Over the past few decades, fault diagnosis of industrial processes has seen significant development. These methods have progressed from mechanism model-based methods to knowledge-based reasoning methods and advanced to data-driven fault diagnosis \cite{ref4}. Mechanism model-based fault diagnosis needs to establish an accurate mathematical model to simulate the behavior of the system and explore the internal operating rules \cite{ref5}. However, it is difficult and time-consuming to establish an accurate mathematical model for the larger and complex modern industrial processes. In contrast, knowledge-based reasoning depends on expert knowledge and experience to identify possible faults. These methods are helpful when the system is too intricate to model accurately using mathematical approaches \cite{ref6}. Nevertheless, the efficacy of these approaches is constrained by the accuracy and comprehensiveness of the knowledge base. Data-driven fault diagnosis that learns fault information and features from large amounts of historical data \cite{ref7, ref8}, has become a critical approach in dealing with complex industrial processes \cite{ref9}.

The core of data-driven fault diagnosis lies in fault data, but gathering fault data from modern industrial systems presents complex attributes. These include high dimensionality \cite{ref10}, nonlinearity \cite{ref11}, noise and outliers \cite{ref12, ref13}, lacking labels \cite{ref14} and occasionally containing zero-samples \cite{ref15} due to the demanding production environment. To address these challenges, researchers have devised a range of methods for data-driven fault diagnosis. For instance, to reduce the dimension of high-dimensional fault data, typical machine learning algorithms such as principal component analysis (PCA) \cite{ref16}, linear discriminant analysis (LDA) \cite{ref17}, and partial least squares (PLS) \cite{ref18} have been applied in fault diagnosis. In addition, Erfani et al. \cite{High} proposed a linear fault-diagnosis method that combines SVM and deep learning. However, the high dimensionality is often coupled with nonlinearity, making it necessary to use nonlinear reduction methods to handle the nonlinear relationships in the data for effective fault diagnosis. Thus, Navi et al. \cite{ref19} extended PCA to kernel principal component analysis (KPCA) for fault diagnosis of nonlinear time-varying systems. Zheng et al. \cite{ref20} proposed a nonlinear fault detection and diagnosis method with the traditional support vector machines (SVM) \cite{ref21}. In recent years, there has been a growing recognition of the superior capabilities of deep learning in extracting deep-level nonlinear information from data. However, the training process of deep neural networks demands a substantial amount of labeled data \cite{refTT, refTT1}, which can be resource-intensive and impractical for small-scale datasets \cite{ref14, ref15}.

Another problem is the presence of noise and outliers in fault samples \cite{ref22} that significantly modifies the distribution of the fault data, thereby affecting the performance of fault diagnosis \cite{ref23}. In order to eliminate the effect of noise, most fault diagnosis methods first learn the characteristics of noise and then suppress noise information. For example, Kai et al. \cite{ref24} proposed a deep noise filtering diagnosis (DNFD) model for accurately and quickly evaluating power transformer faults using noisy vibration signals. Zhang et al. \cite{ref25} proposed an integrated method using deep contractive auto-encoder (EDCAE) for intelligent fault diagnosis of machines in noisy environments. It is important to note that not all data contains noise, and applying noise suppression methods can inadvertently remove the intrinsic information present in noise-free data. As for handling outliers, most research only focuses on handling outliers in fault detection. For example, Xie et al. \cite{ref23} proposed an advanced partial least squares (APLS) based on the data-driven method for handling outlier data in industrial process data. Ferdowsi et al. \cite{ref13} proposed an online outlier identification and removal (OIR) method for nonlinear dynamic systems. Cai et al. \cite{ref26} proposed an intelligent integrated two-stage method that can identify degraded data and mark outlier values simultaneously.

Furthermore, the co-existence of high dimensionality, nonlinearity, noise, and outliers in fault data is common. However, methods addressing individual issues are prevalent, while there are limited approaches capable of addressing the challenges arising from the convergence of these issues. Take the concurrent problems of both high dimensionality and noise as an example, researchers commonly use two strategies: denoising before dimensionality reduction or dimensionality reduction before denoising \cite{ref27, ref28}. While these two-stage approaches combine the advantages of denoising and dimensionality reduction methods, sequential processing can sometimes disrupt the inherent characteristics of the data, resulting in a decline in overall method performance. Therefore, existing data-driven fault diagnosis methods have limited effectiveness due to:

\begin{itemize}
\item In practical scenarios, collecting data for fault samples is challenging because systems are typically not allowed to run to failure. As a result, existing deep-learning methods are not practical for small-scale fault data with limited labeling. On the other hand, machine learning-based methods struggle to handle high-dimensional fault data with nonlinear characteristics.

\item When dealing with concurrent problems, sequential processing approaches inadvertently disrupt the inherent nature of the data. This disruption can lead to the loss of critical information and a decrease in overall method performance, especially when dealing with complex fault data.

\item Current noise-processing methods often rely on assuming the noise distribution or learning its characteristics in advance, followed by noise suppression. However, these methods may inadvertently remove essential fault-related information during the process. Furthermore, when it comes to handling outliers, existing methods are primarily focused on fault detection rather than diagnosis.
\end{itemize}

To address the above problems, a robust unsupervised fault diagnosis method is proposed in this paper. To handle the demand for small-scale fault samples with limited labeling, the proposed method is developed based on conventional machine learning instead of deep learning. Firstly, the high-dimensional data is projected onto a specially designed reduced-dimensional space. Within this space, our goal is to minimize the local scatter among all fault instances while simultaneously maximizing the total scatter. Secondly, the proposed method incorporates nonlinear information related to the faults by using a graph structure learned from the original fault data. Thirdly, recognizing that the impact of outliers and noise can be amplified during the feature extracted process, we introduce the $l_{2,1}$-norm and typicality-aware constraints from a model optimization perspective. The $l_{2,1}$-norm constraint aids in mitigating the influence of noise during the dimension reduction projection. Simultaneously, the typicality-aware constraint harnesses the intrinsic structure of faults to learn the graph adjacency matrix, thereby minimizing the disruptive effects of noise on the overall results. The main contributions of this paper are:

\begin{itemize}
  \item New robust unsupervised fault diagnosis based on a machine learning method is proposed, which includes a nonlinear dimensional reduction model based on graph structure embedding and noise-robust constraints. This method concurrently addresses noise and high-dimensional challenges and is well-aligned with the practical context of fault diagnosis.

  \item The proposed method can effectively address the co-existence of high dimensional, nonlinearity, and noise in fault data. Furthermore, the proposed method is supported by sufficient theoretical proof.

  \item From a model optimization perspective, the proposed method effectively handles noise, making it adaptable to diverse noise types, including outliers and signal noise.
\end{itemize}

\begin{figure*}[h]
\center
    \includegraphics[scale=0.5]{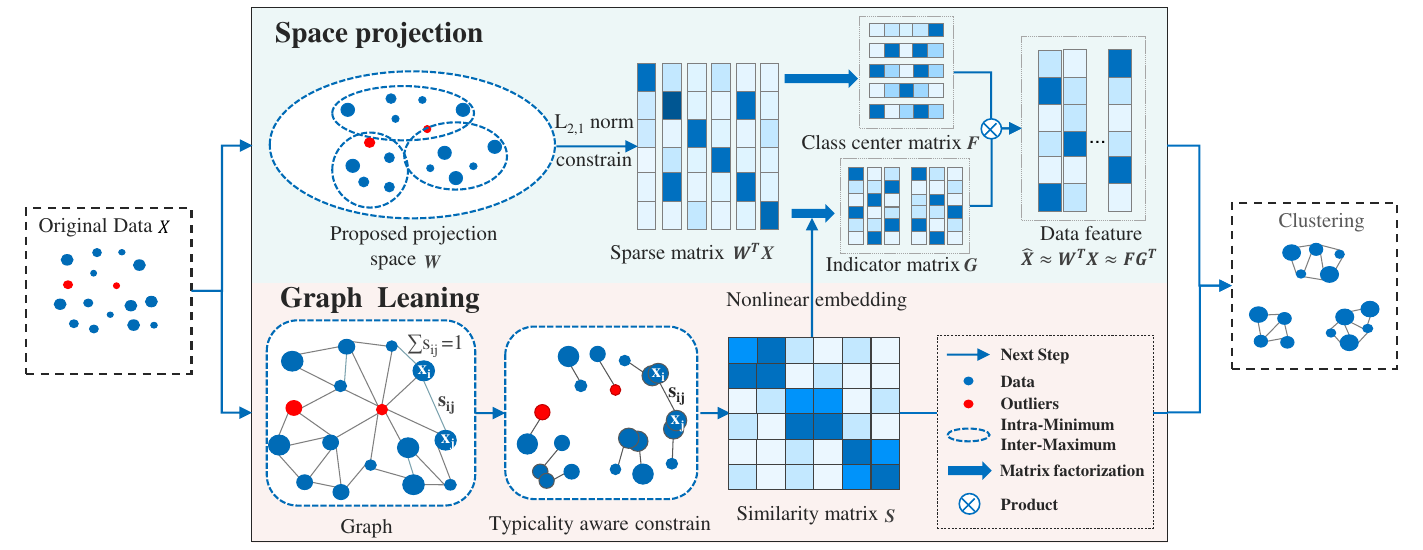}
    \caption{The framework of the proposed unsupervised fault diagnosis method.}\label{FW}
\end{figure*}

\section{Background}
Let $\bm{X} \in \mathbb{R}^{d\times n} $ represent the fault data, where $n$ is the number of faults and $d$ is the dimension of each fault. $ Y \in \mathbb{R}^{1\times n} $ denotes the fault label. The primary objective of dimension reduction is to determine a transformation matrix $\bm{W} \in \mathbb{R}^{d \times m}$ capable of projecting the $d$-dimensional data on an $m$-dimensional space. In the traditional linear discriminant analysis method \cite{refB28}, the core concept involves learning a transformation matrix $W$ to maximize the between-class scatter while minimizing the within-class scatter. Specifically, the total scatter matrix $\bm{S_t}$, between-class scatter matrix $\bm{S_b}$, and within-class scatter matrix $\bm{S_w}$ can be defined as follows:
\begin{equation}
	\begin{aligned}
    & \bm{S_t} = \bm{X}\bm{H}\bm{X}^T\\
    & \bm{S_b} = \bm{X}\bm{H}\bm{G}(\bm{G}^T\bm{G})^{-1}\bm{G}^T\bm{H}\bm{X}^T\\
    & \bm{S_w} = \bm{S_t} - \bm{S_b} = \bm{X}\bm{H}(\bm{I} - \bm{G}(\bm{G}^T\bm{G})^{-1}\bm{G}^T)\bm{H}\bm{X}^T,
    \end{aligned}
\end{equation}
where $\bm{H} \in \mathbb{R}^{n \times n}$ represents the centering matrix, and $\bm{G} \in \mathbb{R}^{n \times c}$ is the label indicator matrix. Here, $c$ is the number of fault classes. In this matrix, $g_{ij} = 1$ indicates that the $i$-th instance belongs to the $j$-th class, while all other elements of $\bm{G}$ are 0. The objective function for linear discriminant analysis is:
\begin{equation}
	\begin{aligned}\label{LDA}
    \max_{\bm{W}} \frac{Tr(\bm{W}^T\bm{S_b}\bm{W})}{Tr(\bm{W}^T\bm{S_w}\bm{W})}.
    \end{aligned}
\end{equation}

\section{Methodology}
This section presents a new unsupervised fault diagnosis method, shown in Figure \ref{FW}. Its main objective is to reduce the dimensional of fault data and extract discriminate features from a small-scale dataset. This is achieved by projecting the fault data on a carefully designed reduced-dimensional space. Subsequently, a nonlinear model captures the intricate geometric structure of the faults. Additionally, to counter the impact of various noise sources, the method incorporates $l_{2,1}$ norm and typicality-aware constraints.

\subsection{Nonlinear Dimensional Reduction}
Let $\bm{W}^T\bm{X}$ represent the projection of $\bm{X}$ on the reduced dimensional space defined by $\bm{W}$. Since obtaining fault-label information is challenging in practical scenarios, the proposed method is designed as an unsupervised approach. This means that the fault labels $\bm{Y}$ are not used in model learning, but only for comparison purposes with the predicted labels $\hat{\bm{Y}} \in \mathbb{R}^{1 \times n}$ generated by our method. Drawing inspiration from matrix factorization (MF), we propose to decompose the features after dimensional reduction into a class center matrix $\bm{F}$ and a class indicator matrix $\bm{G}$ in the reduced dimensional space. Therefore, the indicator matrix $\bm{G}$ becomes a variable that requires optimization. While the center matrix $\bm{H}$ of the original data remains fixed as the mean, the features in the reduced dimensional space should be optimized such that the clustering centers are adaptive, rather than being fixed means. Based on this, we have the following formulation:
 \begin{equation}
	\begin{aligned}\label{eq1}
    &\min_{\bm{W}, \bm{G}, \bm{F}} \Vert\bm{W}^T\bm{X}-\bm{F}\bm{G}^T\Vert_{F}^2\\
     s.t.&\quad \bm{W}^T\bm{W}=\bm{I}, \bm{G}^T\bm{G}=\bm{I},
    \end{aligned}
\end{equation}
where $\bm{F} \in \mathbb{R}^{m \times c}$ and $\bm{G} \in \mathbb{R}^{n \times c}$. Additionally, an important result, referred to as \textit{Theorem 1}, proves that the proposed equation (\ref{eq1}) has the capability to maximize the between-class scatter.

\textit{\textbf{Theorem 1:}} In the projection space defined by $\bm{W}$, we have the following equivalence:
 \begin{equation}
	\begin{aligned}\label{eq4}
    \min_{\bm{F},\bm{G}} \Vert\bm{W}^T\bm{X}-&\bm{F}\bm{G}^T\Vert_{F}^2\\
    &\Leftrightarrow  \min_{\bm{G}} Tr(\bm{W}^T\bm{S_w}\bm{W}).
    \end{aligned}
\end{equation}

\textit{\textbf{Proof:}} Let $\mathcal{F}{\text{right}} = \min \text{Tr}(\bm{W}^T\bm{S_w}\bm{W})$ and $\mathcal{F}{\text{left}} = \min |\bm{W}^T\bm{X} - \bm{F}\bm{G}^T|_{F}^2$. Based on the properties of the matrix norm, we have the following relationship:
 \begin{equation}
	\begin{aligned}\label{eq5}
    \mathcal{F}_{left}& = Tr((\bm{W}^T\bm{X}-\bm{F}\bm{G}^T)(\bm{W}^T\bm{X}-\bm{F}\bm{G}^T)^T)\\
    & = Tr(\bm{W}^T\bm{X}\bm{X}^T\bm{W}-\bm{W}^T\bm{X}\bm{G}\bm{F}^T\\
    &-\bm{F}\bm{G}^T\bm{X}^T\bm{W}+\bm{F}\bm{G}^T\bm{G}\bm{F}^T).
    \end{aligned}
\end{equation}

Differentiating $\mathcal{F}_{\text{left}}$ with respect to $\bm{F}$ and setting it to zero, we can derive the following result:
 \begin{equation}
	\begin{aligned}\label{eq6}
    \bm{F} = \bm{W}^T\bm{X}\bm{G}(\bm{G}^T\bm{G})^{-1}.
    \end{aligned}
\end{equation}

By substituting Eq. (\ref{eq6}) into Eq. (\ref{eq5}), and using the properties of matrix trace $Tr(\bm{A}\bm{B}\bm{C}) = Tr(\bm{B}\bm{C}\bm{A}) = Tr(\bm{C}\bm{B}\bm{A})$, where $\bm{A}, \bm{B}, \bm{C} \in \mathbb{R}^{n \times n}$, we obtain the following expression:
 \begin{equation}
	\begin{aligned}\label{eq7}
    \mathcal{F}_{left}
     = & Tr(\bm{W}^T\bm{X}\bm{X}^T\bm{W}\\&-\bm{W}^T\bm{X}\bm{G}(\bm{G}^T\bm{G})^{-1}\bm{G}^T\bm{X}^T\bm{W})\\
     = & Tr(\bm{W}^T\bm{X}(\bm{I}-\bm{G}(\bm{G}^T\bm{G})^{-1}\bm{G}^T)\bm{X}^T\bm{W},
    \end{aligned}
\end{equation}
where, the $\bm{X}$ refers to the fault instances after centralization. Therefore, we can express it as follows:
 \begin{equation}
	\begin{aligned}\label{eq8}
   \mathcal{F}_{left} \Leftrightarrow &\mathrm{min} Tr(\bm{W}^T\bm{S_w}\bm{W})= \mathcal{F}_{right}.
    \end{aligned}
\end{equation}

Thus, \textit{Theorem 1} is proven.

To incorporate the nonlinear structure and extract discriminate features of faults, the geometric structure is integrated into the model presented in Eq. (\ref{eq1}) through a graph learning approach. In this graph learning model, each fault instance is treated as a point, and the relationships between any two instances are represented as edges. Let $\bm{S}$ denote the adjacency matrix, where$s_{ij}$ is the similarity probability between $x_i$ and $x_j$. $\bm{D_S}$ represents the degree matrix, which can be obtained by diagonalizing $\sum_{j}\bm{s}_{ij}$. One important property of the adjacency matrix is that if there are $k$ connected components, the rank constraint of the normalized Laplacian matrix $\bm{L_S} = \bm{D_S} - (\bm{S}^T - \bm{S})/2 \in \mathbb{R}^{n \times n}$ is given by rank($\bm{L_S}$). To embed the nonlinear geometry of the fault data into Eq. (\ref{eq1}), the Laplacian matrix $\bm{L_S}$ is used to provide valuable guidance for the class indicator matrix. The modified formulation of Eq. (\ref{eq1}) incorporating the Laplacian matrix can be expressed as:
 \begin{equation}
	\begin{aligned}\label{eq9}
    &\min_{\bm{W}, \bm{G}, \bm{F}, \bm{S}} \Vert\bm{W}^T\bm{X}-\bm{F}\bm{G}^T\Vert_{F}^2 +\lambda Tr(\bm{G}^T\bm{L_S}\bm{G})\\
     s.t.&\quad \bm{W}^T\bm{W}=\bm{I}, \bm{G}^T\bm{G}=\bm{I}, \\&\sum_{j}s_{ij} = 1, 0 \leq s_{ij} \leq 1, rank(\bm{L_S})=n-k,
    \end{aligned}
\end{equation}
where $\lambda$ is the parameter for the graph constraint term, the rank constraint for the normalized Laplacian matrix is given by rank($\bm{L_S}$) = $n-k$, which enforces the adjacency matrix to have $k$ connected components.

\subsection{Noise-Robust Constraints}
Although the proposed model enables nonlinear dimensional reduction of high-dimensional fault instances, the proposed method also needs to address the presence of noise and outliers in the original fault data. This is important because the collection environment for fault samples often contains strong noise, and sensors may degrade over time. In the following, we first analyze the impact of noise and outliers on the accuracy of the fault diagnosis model and design specific solution modules to address these challenges via spatial mapping and graph learning.

In the spatial mapping term of Eq. (\ref{eq9}), the $l_2$-norm is used to measure the loss among all the samples. By minimizing the $l_2$-norm, we seek to ensure that instances belonging to the same category are close to each other and have the same distance from their respective center point. However, when there are noisy data or outliers present in the fault data, the distribution of the original faults can be significantly altered. In such cases, using the $l_2$-norm to measure the loss can lead to an amplification of the effect of noisy data or outliers. The $l_2$-norm considers the squared Euclidean distance, which means that larger deviations between the components of a vector will contribute more to the overall magnitude. As a result, noisy data or outliers can have a more pronounced impact on the optimization process, potentially leading to sub-optimal results in terms of fault diagnosis accuracy. In contrast, the $l_1$-norm is less sensitive to noisy data and outliers, since it assigns equal weight to both small and large deviations, effectively reducing the influence of noisy data or outliers. However, the $l_1$ norm constraint typically encourages sparsity of individual elements, as it is sensitive to noise and unrelated features, and pushes their coefficients towards zero in the process of optimization. This can lead to the loss of important information and a decline in model performance.

Therefore, this paper proposes the use of the $l_{2,1}$ norm. The $l_{2,1}$ norm encourages group sparsity, meaning it tends to promote zeroing out entire groups of elements rather than individual elements. This property can make the $l_{2,1}$ norm more robust to noise because it allows for the preservation of relevant groups while effectively suppressing the noise in other groups. By considering the squared Euclidean distance within each group and then summing the square roots across groups, the $l_{2,1}$-norm can provide a balance between individual element sparsity and group-level sparsity. Then Eq. (\ref{eq9}) is rewritten as:
 \begin{equation}
	\begin{aligned}\label{eq10}
    &\min_{\bm{W}, \bm{G}, \bm{F}, \bm{S}} \frac{\Vert\bm{W}^T\bm{X}-\bm{F}\bm{G}^T\Vert_{2,1}}{\Vert \bm{X}^T\bm{W}\Vert_{2,1}} +\lambda Tr(\bm{G}^T\bm{L_S}\bm{G})\\
     s.t.&\quad \bm{W}^T\bm{W}=\bm{I}, \bm{G}^T\bm{G}=\bm{I}, \\&\sum_{j}s_{ij} = 1, 0 \leq s_{ij} \leq 1, rank(\bm{L_S})=n-k,
    \end{aligned}
\end{equation}
where, the term of $||\bm{X}^T\bm{W}||_{2,1}$ is introduced to maximize the global scatter, which seeks to learn robustly and discriminate features for high accuracy fault diagnosis.

On the other hand, in the graph-learning term of Eq. (\ref{eq9}), the constructed adjacency matrix is often worse due to the effect of noisy data and outliers. More specifically, the constraint term of $\sum_{j}s{ij} =1$ implies that the similarity probability $s_{ij}$ can be influenced by other samples than $x_i$ and $x_j$. For instance, if there are $N$ neighbors equidistant from sample $x_i$, the probabilities of these $N$ samples being neighbors of $x_i$ are equal to $1/N$. In this case, if there are a few noisy or outlier data points located closely but far away from the normal data, they may form a separate connected component. As a result, the different normal fault clusters can inevitably merge. This can lead to poor performance in fault diagnosis. To this end, a typicality-aware idea is introduced to describe the similarity probability, it means that the similarity probability $s_{ij}$ is only related to the distance between $x_i$ and $x_j$. Therefore, the objective function of Eq. (\ref{eq10}) is rewritten as follows:
 \begin{equation}
	\begin{aligned}\label{eq11}
		\min_{\bm{W},\bm{F},\bm{G}, \bm{S}} & \quad\frac{\Vert\bm{W}^T\bm{X}-\bm{F}\bm{G}^T\Vert_{2,1}}{\Vert \bm{X}^T\bm{W}\Vert_{2,1}} +\lambda Tr(\bm{G}^T\bm{L_S}\bm{G})\\
		&\quad\quad+\beta\sum_{i=1}^{n}\gamma_i\sum_{j=1}^{n}s_{ij}\\
		 s.t.&\quad \bm{W}^T\bm{W}=\bm{I}, \bm{G}^T\bm{G}=\bm{I}, \\&0\le s_{ij} \le1, rank(\bm{L_S})=n-k,
    \end{aligned}
\end{equation}
where, $\gamma_i \geq 0$ is the balance factor scaled independently for each sample, and $\beta$ is the weight parameter of sample $x_i$ in the adjacency matrix. However, minimizing only $s_{ij}$ in the last term of Equation (\ref{eq11}) can lead to a trivial solution. To address this issue and drawing inspiration from Zhou et al. \cite{ref29}, a typicality-aware graph learning constraint term is designed as follows:
\begin{equation}
	\begin{aligned}\label{eq12}
		\min_{\bm{W},\bm{F},\bm{G}, \bm{S}} & \quad\frac{\Vert\bm{W}^T\bm{X}-\bm{F}\bm{G}^T\Vert_{2,1}}{\Vert \bm{X}^T\bm{W}\Vert_{2,1}} +\lambda Tr(\bm{G}^T\bm{L_S}\bm{G})\\
		&\quad\quad+\beta\sum_{i=1}^{n}\gamma_i\sum_{j=1}^{n}\left(s_{ij}\log s_{ij}-s_{ij}\right)\\
		 s.t.&\quad \bm{W}^T\bm{W}=\bm{I}, \bm{G}^T\bm{G}=\bm{I}, \\&0\le s_{ij} \le1, rank(\bm{L_S})=n-k,
	\end{aligned}
\end{equation}
where, the solution of $s_{ij}\log s_{ij}-s_{ij}$ is a curved surface, which aids in efficiently achieving the optimal solution for the proposed method.

\section{Optimization}
In the proposed objective function, there are four variables that need to be optimized: the center matrix $\bm{F}$, the class indicator matrix $\bm{G}$, the similarity matrix $\bm{S}$ and the projection matrix $\bm{W}$. The optimization of these variables is achieved using fast gradient iterative updating, where three variables are updated and the remaining one is iteratively changed.

\subsection{Fix $\bm{W}$, $\bm{G}$ and $\bm{S}$, update $\mathbf{F}$}
The objective function becomes:
\begin{equation}
	\begin{aligned}\label{eqF}
		\min_{\bm{F}} & \quad\Vert\bm{W}^T\bm{X}-\bm{F}\bm{G}^T\Vert_{2,1},
	\end{aligned}
\end{equation}
where the objective function is constrained by the $l_{2,1}$-norm, which cannot be directly differentiated. We introduce a variable D and  present the following theorem:

\textit{\textbf{Theorem 2:}} Assuming that $\bm{Q}\in\mathbb{R}^{n\times n}$ is a diagonal matrix, with each diagonal entry:
\begin{equation}
 	\begin{aligned}
 		q_{ii}=\frac{1}{(2\|(\bm{W}^T\bm{X}-\bm{F}\bm{G}^T)_i\|_2)^{\frac{1}{2}}},
 	\end{aligned}
\end{equation}
the following equivalence holds:
\begin{equation}
	\begin{aligned}\label{updateF}
		\min_{\bm{F}}  \Vert\bm{W}^T\bm{X}-\bm{F}&\bm{G}^T\Vert_{2,1}\\
		\Leftrightarrow &\min_{\bm{F}}\Vert\left(\bm{W}^T\bm{X}-\bm{F}\bm{G}^T\right)\bm{Q}\Vert_{F}^2.
	\end{aligned}
\end{equation}

To prove Theorem 2, we first need to prove the following lemma.

\textit{\textbf{Lemma 1:}} $\min_{\bm{U}}||\bm{U}||_{2,1}\Leftrightarrow \min_{\bm{U}} Tr(\bm{U}^T\bm{D}\bm{U})$, where $\bm{D}$ is a diagonal matrix with $d_{ii} = \frac{1}{2||\bm{U}_i||_2}$.

\textit{\textbf{Proof:}} For the optimal problem $\min_{\bm{M}}\frac{1}{\Upsilon}||\bm{X}^T\bm{M}-\bm{P}||_{2,1}+||\bm{M}||_{2,1}$. The left term can be rewritten as:
\begin{equation}
	\begin{aligned}\label{updateU}
		\mathcal{F}{\text{left}} =\min_{\bm{U}}  \Vert\bm{U}\Vert_{2,1}, s.t. \bm{Z}\bm{U} = \bm{P},
	\end{aligned}
\end{equation}
where, $\bm{Z} = [\bm{X}^T \quad\Upsilon \bm{I}]\in \mathbb{R}^{n\times m}$, $\bm{I}\in \mathbb{R}^{n\times n}$ is identity matrix. $\bm{U} = \begin{pmatrix}\bm{M} \\ \bm{I} \\ \end{pmatrix}\in \mathbb{R}^{m\times c}$. The Lagrange function of Eq. (\ref{updateU}) can be written as:
\begin{equation}
	\begin{aligned}\label{eq17}
		\bm{L}_1(\bm{U}) = \Vert\bm{U}\Vert_{2,1} - Tr(\Lambda^T(\bm{Z}\bm{U}-\bm{P})),
	\end{aligned}
\end{equation}
where $\Lambda$ is the Lagrange parameter. Let $\frac{\eth\bm{L}_1(\bm{U})}{\eth\bm{U}} = 0$, we have
\begin{equation}
	\begin{aligned}\label{eq177}
		2\bm{D}\bm{U}+\bm{Z}\Lambda^ =0,
	\end{aligned}
\end{equation}

Then, multiply $\bm{Z}\bm{D}^{-1}$ on both side, and substituting $\bm{Z}\bm{U} = \bm{P}$ in it, to get
\begin{equation}
	\begin{aligned}\label{eq178}
		\Lambda = 2(\bm{Z}\bm{D}^{-1}\bm{Z}^T)^{-1}\bm{P}.
	\end{aligned}
\end{equation}

Substituting Eq. (\ref{eq178}) into Eq. (\ref{eq177}), we have $\mathcal{F}{\text{left}} = \bm{D}^{-1}\bm{Z}^T(\bm{Z}\bm{D}^{-1}\bm{A}^T)^{-1}\bm{P}$.

For the right term, it can be rewritten as:
\begin{equation}
	\begin{aligned}\label{eq18}
		\mathcal{F}{\text{right}} =\min_{\bm{U}}  Tr(\bm{U}^T\bm{D}\bm{U}), s.t. \bm{Z}\bm{U} = \bm{P},
	\end{aligned}
\end{equation}
 the Lagrange function of Eq. (\ref{eq18}) can be write as:
\begin{equation}
	\begin{aligned}\label{eq18}
		\bm{L}_2(\bm{U}) = Tr(\bm{U}^T\bm{D}\bm{U})- Tr(\Lambda^T(\bm{Z}\bm{U}-\bm{P})).
	\end{aligned}
\end{equation}

Again, let $\frac{\eth\bm{L}_2(\bm{U})}{\eth\bm{U}} = 0$, and we can derive that $\mathcal{F}{\text{right}} = \bm{D}^{-1}\bm{Z}^T(\bm{Z}\bm{D}^{-1}\bm{A}^T)^{-1}\bm{P} = \mathcal{F}{\text{left}}$. Therefore, Lemma 1 is proved and Theorem 2 is achieved.

According to Theorem 2, minimizing (\ref{eqF}) is equal to minimizing
\begin{equation}
	\begin{aligned}\label{eq19}
		\text{Tr}\left((\bm{W}^T\bm{X}-\bm{F}\bm{G}^T)\bm{D}(\bm{W}^T\bm{X}-\bm{F}\bm{G}^T)^T\right),
	\end{aligned}
\end{equation}
where $\bm{D}\in\mathbb{R}^{n\times n}$ is a diagonal matrix, with each diagonal entry being:
\begin{equation}
	\begin{aligned}\label{eq20}
		d_{ii}=\frac{1}{2\|(\bm{W}^T\bm{X}-\bm{F}\bm{G}^T)_i\|_2}.
	\end{aligned}
\end{equation}

Since the optimal function have the orthogonal constraint that $\bm{W}^T\bm{W}=\bm{I}, \bm{G}^T\bm{G}=\bm{I}$, we expand the above problem as:
\begin{equation}
	\begin{aligned}\label{eq21}
		\min_{\bm{F}}\text{Tr}\left(\bm{F}\bm{G}^T\bm{D}\bm{G}\bm{F}^T\right)-2\text{Tr}\left( \bm{F}\bm{G}^T\bm{D}\bm{X}^T\bm{W} \right).
	\end{aligned}
\end{equation}

Taking the derivative to be zero with respect to $\bm{F}$, we derive the optimal solution:
\begin{equation}
	\begin{aligned}\label{eq22}
		\bm{F}=\bm{W}^T\bm{X}\bm{D}\bm{G}\left(\bm{G}\bm{D}\bm{G}^T\right)^{-1}.
	\end{aligned}
\end{equation}

\subsection{Fix $\bm{W}$, $\bm{S}$ and $\bm{F}$, update $\mathbf{G}$}
To update $\bm{G}$, the objective function Eq. (\ref{eq12}) can be rewritten as:
\begin{equation}
	\begin{aligned}\label{eq23}
			\min_{\bm{G}}\Vert\bm{W}^T\bm{X}-\bm{F}\bm{G}^T\Vert_{2,1} +\lambda Tr(\bm{G}^T\bm{L_S}\bm{G}).
	\end{aligned}
\end{equation}

Using the similar methods as with updating variable $\bm{F}$, we get
\begin{equation}
	\begin{aligned}
			\min_{\bm{G}^T\bm{G}=\bm{I}}\text{Tr}\left(\bm{G}^T\bm{D}\bm{G}\bm{F}^T\bm{F}\right)&-2\text{Tr}\left(\bm{G}^T\bm{D}\bm{X}^T\bm{W} \bm{F}\right)\\
			 &+\lambda\text{Tr}(\bm{G}^T\bm{L_S}\bm{G}),
	\end{aligned}
\end{equation}
which can be rewritten as:
\begin{equation}
	\begin{aligned}\label{QP}
		\min_{\bm{G}^T\bm{G}=\bm{I}}\text{Tr}(\bm{G}^T\bm{R}_1\bm{G})-\text{Tr}\left(\bm{G}^T\bm{D}\bm{R}_2\right),
	\end{aligned}
\end{equation}
where $\bm{R}_1 = \lambda\bm{L_S}$ and $\bm{R}_2=\bm{X}^T\bm{W} \bm{F}-\frac{1}{2}\bm{G}\bm{F}^T\bm{F}$. To make matrix positive definite, the problem (\ref{QP}) can be further refined as:
\begin{equation}
	\begin{aligned}\label{MQP}
		\max_{\bm{G}^T\bm{G}=\bm{I}}\text{Tr}(\bm{G}^T(\sigma_{max}\bm{I}-\bm{R}_1)\bm{G})+\text{Tr}\left(\bm{G}^T\bm{D}\bm{R}_2\right),
	\end{aligned}
\end{equation}
where $\sigma_{max}$ is the maximal eigenvalue of $\mathbf{R}_1$, problem (\ref{MQP}) is the well-known quadratic problem on stiefel manifold (QPSM), which can be effectively solved by generalized power iteration (GPI) method \cite{ref30}.

\subsection{Fix $\bm{W}$, $\bm{F}$ and $\bm{G}$, updating $\mathbf{S}$}
To update $\bm{S}$, the objective function Eq. (\ref{eq12}) can be rewritten as:
\begin{equation}
	\begin{aligned}\label{updateS}
		\min_{\bm{S}} & \quad Tr(\bm{G}^T\bm{L_S}\bm{G})+\beta\sum_{i=1}^{n}\gamma_i\sum_{j=1}^{n}\left(s_{ij}\log s_{ij}-s_{ij}\right)\\
		s.t.&\quad  0\le s_{ij} \le1, rank(\bm{L_S})=n-k,
	\end{aligned}
\end{equation}
which can be reformed as:
\begin{equation}
	\begin{aligned}\label{eq31}
		\min_{\bm{S}} & \sum_{i=1}^{n}\sum_{j=1}^{n}\left(\|\bm{g}_i-\bm{g}_j\|_2^2s_{ij}\right)+\beta\sum_{i=1}^{n}\gamma_i\sum_{j=1}^{n}\left(s_{ij}\log s_{ij}-s_{ij}\right)\\
		s.t.&  0\le s_{ij} \le1, rank(\bm{L_S})=n-k.
	\end{aligned}
\end{equation}

Let $d_{ij}^2=\|\bm{g}_i-\bm{g}_j\|_2^2$, to give
\begin{equation}
	\begin{aligned}\label{dij}
		\mathcal{L}_{ASW} &=\sum_{i=1}^{n}\sum_{j=1}^{n}\left(d_{ij}^2s_{ij}\right)+\beta\sum_{i=1}^{n}\gamma_i\sum_{j=1}^{n}\left(s_{ij}\log s_{ij}-s_{ij}\right)\\
		s.t.&  \quad0\le s_{ij} \le1,  rank(\bm{L_S})=n-k.
	\end{aligned}
\end{equation}

The deviation of Eq. (\ref{dij}) with respect to $s_{ij}$ yields

\begin{equation}
	\begin{aligned}\label{pdij}
		\frac{\partial\mathcal{L}_{ASW}}{\partial s_{ij}}=d_{ij}^2+\gamma_i\log s_{ij}.
	\end{aligned}
\end{equation}

Let $\frac{\partial\mathcal{L}_{ASW}}{\partial s_{ij}}=0$, then we achieve:
\begin{equation}
	\begin{aligned}\label{sij}
		 s_{ij}=\exp\left(-\frac{d_{ij}^2}{\gamma_i}\right).
	\end{aligned}
\end{equation}

\subsection{Fix $\bm{F}$, $\bm{G}$ and $\bm{S}$, update $\mathbf{W}$}
To update $\bm{W}$, the objective function Eq. (\ref{eq12}) can be rewritten as:
\begin{equation}
	\begin{aligned}\label{updateW}
		\min_{\bm{W}^T\bm{W}=\bm{I}} & \quad\frac{\Vert\bm{W}^T\bm{X}-\bm{F}\bm{G}^T\Vert_{2,1}}{\Vert \bm{X}^T\bm{W}\Vert_{2,1}}.
	\end{aligned}
\end{equation}

Substituting $\bm{F}=\bm{W}^T\bm{X}\bm{D}\bm{G}\left(\bm{G}\bm{D}\bm{G}^T\right)^{-1}$ into Eq. (\ref{updateW}), then we have
\begin{equation}
	\begin{aligned}\label{FupdateW1}
		\min_{\bm{W}^T\bm{W}=\bm{I}} & \frac{\Vert\bm{W}^T\left(\bm{X}-\bm{X}\bm{D}\bm{G}\left(\bm{G}\bm{D}\bm{G}^T\right)^{-1}\bm{G}^T\right)\Vert_{2,1}}{\Vert \bm{X}^T\bm{W}\Vert_{2,1}}.
	\end{aligned}
\end{equation}

Let $\bm{U}=\bm{X}\bm{D}\bm{G}\left(\bm{G}\bm{D}\bm{G}^T\right)^{-1}\bm{G}^T$, then (\ref{FupdateW1}) becomes
\begin{equation}
	\begin{aligned}\label{L24}
		\min_{\bm{W}^T\bm{W}=\bm{I}} & \quad\frac{\Vert\bm{W}^T(\bm{X}-\bm{U})\Vert_{2,1}}{\Vert \bm{X}^T\bm{W}\Vert_{2,1}}=\frac{\sum_{i=1}^{n}\|\bm{W}^T(\bm{x}_i-\bm{\mu}_i)\|_2}{\sum_{i=1}^{n}\|\bm{W}^T\bm{x}_i\|_2},
	\end{aligned}
\end{equation}

For solving problems in which both the numerator and the denominator contain the $l_{21}$-norm, Nie et al. \cite{refTT} proposed a non-greedy fast algorithm. Therefore, we can solve the following problem:
\begin{equation}
	\begin{aligned}\label{sL25}
		\min_{\bm{W}^T\bm{W}=\bm{I}} & \quad\sum_{i=1}^{n}p_i\|\bm{W}^T(\bm{x}_i-\bm{u}_i)\|_2^2-\xi\sum_{i=1}^{n}\bm{\mu}\|\bm{W}^T\bm{x}_i\|_2^2,
	\end{aligned}
\end{equation}
where $\xi=\frac{\sum_{i=1}^{n}\|\bm{W}^T(\bm{x}_i-\bm{\mu}_i)\|_2}{\sum_{i=1}^{n}\|\bm{W}^T\bm{x}_i\|_2}$,  $p_i=\frac{1}{2\|\bm{W}^T(\bm{x}_i-\bm{u}_i)\|_2}$, and $\bm{\mu}$ is defined as:
\begin{equation}\label{mu}
\bm{\mu}=	\left\{
	\begin{array}{lr}
		\frac{\bm{W}^T\bm{x}_i}{\|\bm{W}^T\bm{x}_i\|_2} &, if \|\bm{W}^T\bm{x}_i\|_2\ne 0,  \\
		0 &, if \|\bm{W}^T\bm{x}_i\|_2= 0.
	\end{array}
	\right.
\end{equation}

Let
\begin{equation}
	\begin{aligned}\label{26}
		&\bm{A}=\sum_{i=1}^{n}p_i(\bm{x}_i-\bm{\mu}_i)(\bm{x}_i-\bm{\mu}_i)^T,\\
		&\bm{B}=\sum_{i=1}^{n}\bm{x}_i\bm{\mu}_i^T\in\mathbb{R}^{d\times m},
	\end{aligned}
\end{equation}

Eq. (\ref{sL25}) can be rewritten using matrices to give:

\begin{equation}
	\begin{aligned}\label{27}
		\min_{\bm{W}^T\bm{W}=\bm{I}} & \text{Tr}(\bm{W}^T\bm{A}\bm{W})-2\text{Tr}(\bm{W}^T\bm{B}).
	\end{aligned}
\end{equation}

To make the matrix $\bm{W}^T\bm{A}\bm{W}$ be positive definite, the optimization problem can be defined as:
\begin{equation}
	\begin{aligned}\label{28}
		\min_{\bm{W}^T\bm{W}=\bm{I}} & \text{Tr}(\bm{W}^T\tilde{\bm{A}}\bm{W})-2\text{Tr}(\bm{W}^T\bm{B}),
	\end{aligned}
\end{equation}
where, $\tilde{\bm{A}} = \sigma_{\max}\bm{I}-\sum_{i=1}^{n}p_i(\bm{x}_i-\bm{\mu}_i)(\bm{x}_i-\bm{\mu}_i)^T$, $\sigma_{\max}$ is the maximum eigenvalue of matrix $\tilde{\bm{A}}$. It can be effectively solved by the GPI method. The complete algorithm is summarized as Algorithm 1.

\begin{algorithm}[H]
\caption{Algorithm for Solving Problem (\ref{eq12})}\label{alg:alg1}
\begin{algorithmic}
\STATE
\STATE \textbf{Input:} Fault data matrix $\bm{X}$, parameter $\lambda$ and $\beta$ and \\ \hspace{1cm} the projection dimension $m$.
\STATE \hspace{0.3cm} \textbf{Initialization:} $\bm{W}^1, \bm{G}^1$ and $\bm{F}^1$ as a random matrix, \\ \hspace{0.3cm} respectively.Initialize the adjacency matrix \\ \hspace{0.3cm} $S^1 = [s_{ij}]_{n\times n}$ by data matrix $\bm{X}$, $t = 1$.
\STATE \hspace{0.8cm} \textbf{while} not converged \textbf{Do}
\STATE \hspace{1.1cm} \textbf{1.} Update $\bm{F}^{t+1}$ by Eq. (\ref{eq22});
\STATE \hspace{1.1cm} \textbf{2.} Calculate $\bm{L_S}^{t+1} = \bm{D_S} - \frac{\bm{S}^T - \bm{S}}{2}$, where degree \\\hspace{1.4cm} matrix $\bm{D}$ is a diagonal matrix with an $i$-th \\\hspace{1.4cm} diagonal element of $\sum_{j=1}\frac{s_{ji+s_{ij}}}{2}$;
\STATE \hspace{1.1cm} \textbf{3.} Calculate $\bm{R}_1=\lambda\bm{L_S}$,\\ \hspace{3cm} $\bm{R}_2=\bm{X}^T\bm{W} \bm{F}-\frac{1}{2}\bm{G}\bm{F}^T\bm{F}$;
\STATE \hspace{1.1cm} \textbf{4.} Update $\bm{G}^{t+1}$ by using GPI with $\bm{R}_1$ and $\bm{R}_2$;
\STATE \hspace{1.1cm} \textbf{5.} Update $\bm{S}^{t+1}$ by Eq. (\ref{sij});
\STATE \hspace{1.1cm} \textbf{6.} Calculate $\bm{\mu}$ according to Eq. (\ref{mu}) with $\bm{W}^{t}$;
\STATE \hspace{1.1cm} \textbf{7.} Calculate $\bm{A}=\sum_{i=1}^{n}p_i(\bm{x}_i-\bm{\mu}_i)(\bm{x}_i-\bm{\mu}_i)^T$,\\ \hspace{1.4cm} and the eigenvalue of maximum$\sigma_{\max}$;
\STATE \hspace{1.1cm} \textbf{8.} Calculate $\tilde{\bm{A}}=\sigma_{\max}\bm{I}-\bm{A}$,\\
	 \hspace{3cm} $\bm{B}=\sum_{i=1}^{n}\bm{x}_i\bm{\mu}_i^T$;
\STATE \hspace{1.1cm} \textbf{9.} Update $\bm{W}^{t+1}$ by using GPI with $\bm{A}$ and $\bm{B}$;
\STATE \hspace{1.1cm} \textbf{10.} $t = t+1$;
\STATE \hspace{0.8cm} \textbf{end while}
\STATE \textbf{Output:} The projection matrix $\bm{W}^{t+1}$ and indicator \\ \hspace{1.2cm} matrix $\bm{G}^{t+1}$
\end{algorithmic}
\label{alg1}
\end{algorithm}

\section{Convergence Analysis}
The convergence criterion in an algorithm states that the ratio of the difference between the objective function values of consecutive iterations should be less than or equal to a small constant. It can be formulated as $(obj^{t+1} - obj^{t})/obj^(t) \leq \epsilon$, where $obj$ is the objective function values and $\epsilon$ is a small constant. The proposed objective function (\ref{eq12}) is an optimization problem with fractions, which the convergence analysis of the optimal algorithm can be proved from the following two theorems.

\textit{\textbf{Theorem 3:}} The update of matrix $\bm{F}$, $\bm{G}$ and $\bm{S}$ in Algorithm 1 will result in a decrease in the objective function value of problem (\ref{eq12}) during each iteration until convergence.

\textit{\textbf{Theorem 4:}} The update of matrix $\bm{W}$ in Algorithm 1 will result in a decrease in the objective function value of the problem (\ref{eq12}) during each iteration until convergence.

To prove the above theorems, we first introduce \textit{\textbf{Lemma 2:}} For any nonzero vectors $\bm{a}$ and $\bm{b}$, we obtain the following inequality:
\begin{equation}
	\begin{aligned}\label{C1}
		\|\bm{a}\|_{2}-\frac{\|\bm{a}\|_{2}^{2}}{2\|\bm{b}\|_{2}} \leq\|\bm{b}\|_{2}-\frac{\|\bm{b}\|_{2}^{2}}{2\|\bm{b}\|_{2}}.
	\end{aligned}
\end{equation}

\textit{\textbf{Proof of Lemma 2:}} The following inequality is satisfied obviously for any nonzero vectors $\bm{a}$ and $\bm{b}$:
\begin{equation}
	\begin{aligned}\label{C2}
		(\|\bm{a}\|_{2}-\|\bm{b}\|_{2})^{2} \geq 0,
	\end{aligned}
\end{equation}
expand $(\|\bm{a}\|_{2}-\|\bm{b}\|_{2})^{2}$ and divide both sides of the equation by $2\|\bm{b}\|_{2}$, we get
\begin{equation}
	\begin{aligned}\label{C3}
		\frac{\|\bm{a}\|_{2}^{2}}{2\|\bm{b}\|_{2}}-\|\bm{a}\|_{2}+\frac{\|\bm{b}\|_{2}^{2}}{2\|\bm{b}\|_{2}} \geq 0,
	\end{aligned}
\end{equation}
where $\frac{\|\bm{b}\|_{2}^{2}}{2\|\bm{b}\|_{2}}=\frac{2\|\bm{b}\|_{2}^{2}-\|\bm{b}\|_{2}^{2}}{2\|\bm{b}\|_{2}}=\|\bm{b}\|_{2}-\frac{\|\bm{b}\|_{2}^{2}}{2\|\bm{b}\|_{2}}$. Thus, Lemma 1 has been proved.

\textit{\textbf{Proof of Theorem 3:}}  In updating $\bm{F}$, $\bm{G}$ and $\bm{S}$ , the optimization problem converts to
\begin{equation}
	\begin{aligned}\label{C4}
		\min _{\bm{F}, \bm{G}, \bm{S}}||\bm{W}^{T} \bm{X}-\bm{F} \bm{G}^{T}||_{2,1}+\lambda Tr(\bm{G}^T\bm{L_S}\bm{G}),
	\end{aligned}
\end{equation}
which is equivalent to the following optimization problem:
\begin{equation}
	\begin{aligned}\label{C5}
		\min _{\bm{F}, \bm{G}, \bm{S}}|| (\bm{W}^{T} \bm{X}-\bm{F} \bm{G}^{T}) \bm{Q}||_{F}^{2 }+\lambda Tr(\bm{G}^T\bm{L_S}\bm{G}),
	\end{aligned}
\end{equation}
where $\bm{Q}$ $\in$ $\mathbb{R}$$^{n \times n}$ is a diagonal matrix and $q_{i i}=\frac{1}{\left(2\left\|\left(\bm{W}^{T} \bm{X}-\bm{F} \bm{G}^{T}\right)_{i}\right\|_{2}\right)^{\frac{1}{2}}}$. Thus, the optimization problem (\ref{C5}) converts to
\begin{equation}
	\begin{aligned}\label{C6}
		\sum_{i=1}^{n} \frac{\left\|\left(\bm{W}^{T} \bm{X}-\bm{F} \bm{G}^{T}\right)_{i}\right\|_{2}^{2}}{2\left\|\left(\bm{W}^{T} \bm{X}-\bm{F} \bm{G}^{T}\right)_{i}\right\|_{2}}+\lambda \operatorname{Tr}\left(\bm{G}^{T} \bm{L_{S}} \bm{G}\right).
	\end{aligned}
\end{equation}

Suppose that $\bm{F}$,$\bm{G}$,$\bm{S}$ are updated by $\hat{\bm{F}}$,$\hat{\bm{G}}$,$\hat{\bm{S}}$, respectively. When $\bm{W}$,$\bm{Q}$ is fixed at previous iteration, we have
\begin{equation}
	\begin{aligned}\label{C7}
		&\sum_{i=1}^{n} \frac{\left\|\left(\bm{W}^{T} \bm{X}-\hat{\bm{F}} \hat{\bm{G}}^{T}\right)_{i}\right\|_{2}^{2}}{2\left\|\left(\bm{W}^{T} \bm{X}-\bm{F} \bm{G}^{T}\right)_{i}\right\|_{2}}+\lambda \operatorname{Tr}\left(\hat{\bm{G}}^{T} \hat{\bm{L_S}} \hat{\bm{G}}\right)\\&\leq \sum_{i=1}^{n} \frac{\left\|\left(\bm{W}^{T} \bm{X}-\bm{F} \bm{G}^{T}\right)_{i}\right\|_{2}^{2}}{2\left\|\left(\bm{W}^{T} \bm{X}-\bm{F} \bm{G}^{T}\right)_{i}\right\|_{2}}+\lambda \operatorname{Tr}\left(\bm{G}^{T} \bm{L_{S}} \bm{G}\right).
	\end{aligned}
\end{equation}

According to Lemma 2, we have
\begin{equation}
	\begin{aligned}\label{C8}
		&\left\|\left(\bm{W}^{T} X-\hat{\bm{F}} \hat{\bm{G}}^{T}\right)_{i}\right\|_{2}-\frac{\left\|\left(\bm{W}^{T} \bm{X}-\hat{\bm{F}} \hat{\bm{G}}^{T}\right)_{i}\right\|_{2}^{2}}{2\left\|\left(\bm{W}^{T} \bm{X}-\bm{F} \bm{G}^{T}\right)_{i}\right\|_{2}}\\&\leq\left\|\left(\bm{W}^{T}\bm{X}-\bm{F} \bm{G}^{T}\right)_{i}\right\|_{2}-\frac{\left\|\left(\bm{W}^{T} \bm{X}-\bm{F} \bm{G}^{T}\right)_{i}\right\|_{2}^{2}}{2\left\|\left(\bm{W}^{T} \bm{X}-\bm{F} \bm{G}^{T}\right)_{i}\right\|_{2}}.
	\end{aligned}
\end{equation}

Summing (3) and (4), and substituting $\sum_{i=1}^{n}\left\|\cdot_{i}\right\|_{2}=\left\|\cdot\right\|_{2,1}$, we hold
\begin{equation}
	\begin{aligned}\label{C9}
		&||\bm{W}^{T} \bm{X}-\hat{\bm{F}} \hat{\bm{G}}^{T}||_{2,1}+\lambda \operatorname{Tr}(\hat{\bm{G}}^{\bm{T}} \hat{\bm{L_S}} \hat{\bm{G}})\\&\leq ||\bm{W}^{T} \bm{X}-\bm{F} \bm{G}^{T}||_{2,1}+\lambda \operatorname{Tr}(\bm{G}^{T} \bm{L_S} \bm{G}).
	\end{aligned}
\end{equation}

Therefore, Theorem 3 is proved.

\textit{\textbf{Proof of Theorem 4:}} In updating $\bm{W}$, the optimization problem converts to
\begin{equation}
	\begin{aligned}\label{C10}
		&\min _{\bm{W}} \frac{||\bm{W}^{T} \bm{X}-\bm{F} \bm{G}^{T}||_{2,1}}{||\bm{X}^{T} \bm{W}||_{2,1}}\\
&s.t. \bm{W}^T\bm{W} = \bm{I}.
	\end{aligned}
\end{equation}

Since we have $\bm{F}=\bm{W}^{T} \bm{X} \bm{D} \bm{G}(\bm{G}^{T} \bm{D} \bm{G})^{-1}$, let $\bm{U}=\bm{X} \bm{D} \bm{G}(\bm{G}^{T} \bm{D} \bm{G}t)^{-1} \bm{G}^{T}$, the optimization problem (\ref{C10}) converts to
\begin{equation}
	\begin{aligned}\label{C11}
		\min _{\bm{W}} \frac{||\bm{W}^{T} (\bm{X}-\bm{U}) ||_{2,1}}{||\bm{X}^{T} \bm{W}||_{2,1}}
&=\min _{\bm{W}}\frac{\sum_{i=1}^{n}||\bm{W}^{T}(\bm{x}_{i}-\bm{u}_{i})||_{2}}{\sum_{i=1}^{n}||\bm{W}^{T} \bm{x}_{i}||_{2}}\\
&s.t. \bm{W}^T\bm{W} = \bm{I}.
	\end{aligned}
\end{equation}

In the process of solving $\bm{W}$, we compute $W^{t+1}= \sum_{i=1}^{n} \bm{p}_{i}||\bm{W}^{T}(\bm{x}_{i}-\bm{u}_{i})|  |_{2}^{2}-\lambda^{t} \sum_{i=1}^{n} \mu_{i}^{T} \bm{W}^{T} \bm{x}_{i}$, where $\lambda^{t}=\frac{\sum_{i=1}^{n}||(\bm{W}^{t})^{T}(\bm{x}_{i}-\bm{u}_{i})||_{2}}{\sum_{i=1}^{n}||(\bm{W}^{t})^{T} \bm{x}_{i}||_{2}}$, $\bm{p}_{i}=\frac{1}{2||\bm{W}^{T}(\bm{x}_{i}-\bm{u}_{i})||_{2}}$. Therefore, for any $\bm{W}$ which satisfies $\bm{W}^{T}\bm{W}=\bm{I}$, the following inequality satisfies:
\begin{equation}
	\begin{aligned}\label{C12}
&\sum_{i=1}^{n} \bm{p}_{i}||(\bm{W}^{t+1})^{T}(\bm{x}_{i}-\bm{u}_{i})||_{2}^{2}-\lambda^{t} \sum_{i=1}^{n} \mu_{i}^{T}(\bm{W}^{t+1})^{T} \bm{x}_{i}\\
&\leq \sum_{i=1}^{n} \bm{p}_{i}||(\bm{W}^{t})^{T}(\bm{x}_{i}-\bm{u}_{i})||_{2}^{2}-\lambda^{t} \sum_{i=1}^{n} \mu_{i}^{T}(\bm{W}^{t})^{T} \bm{x}_{i}.
	\end{aligned}
\end{equation}

Since the function $\sqrt{\cdot}$ and $|\cdot|$ are concave and convex functions respectively, we obtain the following two inequalities according to the definition of the supergradient:
\begin{equation}
	\begin{aligned}\label{C13}
&\sum_{i=1}^{n}(||(\bm{W}^{t+1})^{T}(\bm{x}_{i}-\bm{u}_{i})||_{2}-||(\bm{W}^{t})^{T}(\bm{x}_{i}-\bm{u}_{i})||_{2} \\
& +\bm{\mu}_{i}^{T}(\bm{W}^{t+1})^{T} \bm{x}_{i}-\bm{\mu}_{i}^{T}(\bm{W}^{t})^{T} \bm{x}_{i})\\
&\leq\sum_{i=1}^{n}(\bm{p}_{i}||(\bm{W}^{t+1})^{T}(\bm{x}_{i}-\bm{u}_{i})||_{2}^{2}-\bm{p}_{i}||(\bm{W}^{t})^{T}(\bm{x}_{i}-\bm{u}_{i})||_{2}^{2} \\
& +||(\bm{W}^{t+1})^{T} \bm{x}_{i}||_{2}-||(\bm{W}^{t})^{T} \bm{x}_{i}||_{2}).
	\end{aligned}
\end{equation}

Substituting the inequality (\ref{C12}) into (\ref{C13}), we have
\begin{equation}
	\begin{aligned}\label{C14}
&\sum_{i=1}^{n}(||(\bm{W}^{t+1})^{T}(\bm{x}_{i}-\bm{u}_{i})||_{2}-||(\bm{W}^{t})^{T}(\bm{x}_{i}-\bm{u}_{i})||_{2}  \\
&  +\bm{\mu}_{i}^{T}(\bm{W}^{t+1})^{T} \bm{x}_{i}-\bm{\mu}_{i}^{T}(\bm{W}^{t})^{T} \bm{x}_{i})\\
&\leq \lambda^{t} \sum_{i=1}^{n}(\bm{\mu}_{i}^{T}(\bm{W}^{t+1})^{T} \bm{x}_{i}-\bm{\mu}_{i}^{T}(\bm{W}^{t})^{T} \bm{x}_{i})\\
& +\sum_{i=1}^{n}(||(\bm{W}^{t+1})^{T} \bm{x}_{i}||_{2}-||(\bm{W}^{t})^{T} \bm{x}_{i} ||_{2}).
	\end{aligned}
\end{equation}

Then, put $\lambda^{t}=\frac{\sum_{i=1}^{n}||(\bm{W}^{t})^{T}(\bm{x}_{i}-\bm{u}_{i})||_{2}}{\sum_{i=1}^{n}||(\bm{W}^{t})^{T} \bm{x}_{i}||_{2}}$ and $\lambda^{t+1}=\frac{\sum_{i=1}^{n}||(\bm{W}^{t+1})^{T}(\bm{x}_{i}-\bm{u}_{i})||_{2}}{\sum_{i=1}^{n}||(\bm{W}^{t+1})^{T} \bm{x}_{i}||_{2}}$ into inequality (\ref{C14}) and simplify, we have
\begin{equation}
	\begin{aligned}\label{C15}
&(\lambda^{t+1}-\lambda^{t}) \sum_{i=1}^{n}||(\bm{W}^{t+1})^{T} \bm{x}_{i}||_{2} \leq 0.
	\end{aligned}
\end{equation}

Note that $||(\bm{W}^{t+1})^{T} \bm{x}_{i}||_{2}\ge 0$ is always satisfied. So inequality (\ref{C15}) is only satisfies when $\lambda^{t+1} \leq \lambda^{t}$. Therefore, we get the following inequality:
\begin{equation}
	\begin{aligned}\label{C16}
\frac{\sum_{i=1}^{n}||(\bm{W}^{t+1})^{T}(\bm{x}_{i}-\bm{u}_{i})||_{2}}{\sum_{i=1}^{n}||(\bm{W}^{t+1})^{T} \bm{x}_{i}||_{2}} \leq \frac{\sum_{i=1}^{n}||(\bm{W}^{t})^{T}(\bm{x}_{i}-\bm{u}_{i})||_{2}}{\sum_{i=1}^{n}||(W^{t})^{T} \bm{x}_{i}||_{2}}.
	\end{aligned}
\end{equation}

Thus, Theorem 4 is proved.

\section{Experiment and Discussion}
In the experiment section, we seek to answer the following questions:

1. Can the proposed method achieve better fault diagnosis performance for high-dimensional nonlinear data?

2. What proportion of data containing outliers and noise can be handled by the proposed method? and how does it compare with existing methods?

3. How do the hyper-parameters influence the proposed method?

\begin{table}[h!]
  \begin{center}
    \caption{The state presentation of two datasets.}\label{Data1}
    \begin{tabular}{c|c}
    \hline
    \textbf{Fault ID}  &\textbf{Fault State}\\ \hline
    \multicolumn{2}{c}{TE process Dataset}\\  \hline
      T.1   &A/C feed ratio, B composition constant\\
      T.6   & A feed loss\\
      T.14   &Reactor cooling water valve\\
      T.2    &B composition, A/C ration constant\\
      T.3   &D feed temperature\\
      T.5   & Condenser cooling water temperature\\
      T.4  & Reactor cooling water inlet temperature\\
      T.7 & C header pressure loss\\
      T.10  & Temperature of input is changed\\  \hline
    \multicolumn{2}{c}{HSM Dataset}\\ \hline
      S.1  & Roller swing \\
      S.2  & Roller stuck\\
      S.3  & Over current \\
      S.4  & Squeaking\\
      S.5  & Base deformation\\ \hline
    \end{tabular}
  \end{center}
\end{table}
\subsection{Datasets and Evaluation Metrics}
To verify the fault diagnosis performance of the proposed method for high-dimension data and outliers, this section conducted a series of comparison experiments on two datasets, including the simulated Tennessee-Eastman (TE) process \cite{ref33} and one from the real hot rolling of steel process.

\textit{TE process description:} The TE process dataset has been extensively used to validate the effectiveness of fault diagnosis methods, making it crucial for fairly presenting the proposed fault diagnosis method. The dataset consists of a total of 52 variables, which are further classified into 11 manipulated variables and 41 measured variables. The TE process dataset comprises 21 distinct classes of faults. The training set of this dataset consists of 480 samples for each fault class, while the testing set contains 960 samples for each class. Since the proposed method in this paper is an unsupervised clustering method, there is no need to pretrain the model using data. Therefore, only the training data from the TE process dataset was used in the experiment. Specifically, 480 samples from each class were used for unsupervised fault clustering. The diagnostic task in this section follows the experiments of Feng et al. \cite{ref34}, where three fault classes were selected for classification out of a total of 21 classes. Table \ref{Data1} provides a detailed overview of the fault state used, while Table \ref{Data2} is the experimental group used.

\begin{figure}[htbp]
\centering
\includegraphics[scale=0.35]{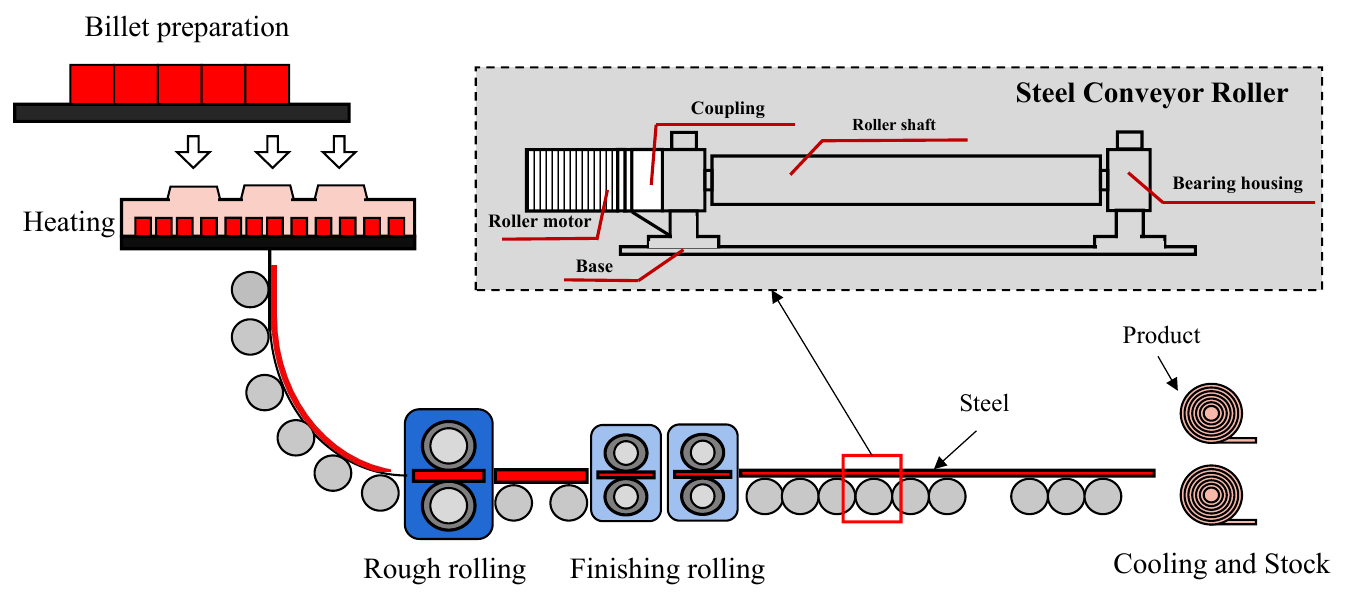}
\caption{The process of steel hot rolling.}\label{CC}
\end{figure}
\textit{HSM dataset description:} Hot steel milling is one of the important processes in steel manufacturing, which involves five key steps: billet preparation, heating, rough rolling, finishing rolling, and cooling. As shown in Fig. \ref{CC}, the steel conveyor rollers are vital equipment that span the entire steel rolling process, and their function is to transfer steel to the next stage. The failure of the steel conveyor rollers can result in steel billet deformation, surface defects, as well as production downtime, and increased costs. This dataset was collected from a real industry of steel hot rolling with a total of 294 rolling transfer rollers. The running status of each roller is recorded by its motor current signal and finally transferred into digital data. Since the sampling frequency is 2 Hz, there are 7200 data points in one hour. Additionally, the environment is harsh, with high temperatures, spray condensation, and oil lubrication, which often introduce strong noise into the data. Therefore, accurate fault diagnosis under high-dimensional and noisy data conditions is of great significance. Over a period of six months, we collected a total of five categories of fault data, as shown in Table \ref{Data1}. Due to the randomness and uncertainty of fault occurrences, the number of fault samples collected in this dataset is limited. As is shown in Table \ref{Data2}, the sample dimension was set to 1000 using a sliding window approach, and 20 samples were collected for each class.

\begin{table}[h!]
  \begin{center}
    \caption{Introduction of experimental data properties.}\label{Data2}
    \begin{tabular}{c|c|c|c}
    \hline
    \textbf{Groups} & \textbf{Fault ID} & \textbf{Dimension}& \textbf{Number of samples} \\ \hline
      1 & T.1, T.6, T.14 & 52& 1440 (480 $\times$ 3)  \\
      2 & T.2, T.3, T.5 & 52 &  1440 (480 $\times$ 3)   \\
      3 & T.4, T.7, T.10 & 52 & 1440 (480 $\times$ 3) \\
      4 &S.1, S.2, S.3, S.4, S.5 &1000 &100 (20 $\times$ 5)\\ \hline
    \end{tabular}
  \end{center}
\end{table}

\begin{table*}[h!]
  \begin{center}
    \caption{ACC of the three groups of TE process dataset.}\label{result1acc}
    \resizebox{\linewidth}{!}{
    \begin{tabular}{cccccccccccccc}
    \hline
    \textbf{Groups}    & \textbf{Dim}     &\textbf{K-means}            &\textbf{Feng (LSVM)}        &\textbf{Feng (RF)}      &\textbf{Feng (NB)}   &\textbf{t-SNE} &\textbf{kPCA} &\textbf{LLE} &\textbf{PCA}  &\textbf{AE}&\textbf{Un-LDA}  &\textbf{Ours-nGraph} &\textbf{Ours}  \\
     \hline
       \multirow{3}{*}{1}  & 5&\multirow{3}{*}{71.81}&\multirow{3}{*}{58.68}&\multirow{3}{*}{88.40}&\multirow{3}{*}{80.27} &{\color{red}\textbf{94.51}} &77.92  &83.54  &92.01 &71.11 &76.81 & 87.15  &{\color{blue}\textbf{93.96}}     \\
                           & 10         & & & & &{\color{blue}\textbf{94.86}}   &78.06   &87.43   &92.98    &61.74    &67.57   &92.47   &{\color{red}\textbf{97.22}}     \\
                           & 25         & & & & &94.58   &77.99   &71.81   &89.61    &90.28    &84.72   &{\color{blue}\textbf{96.78}}   &{\color{red}\textbf{97.85}}     \\ \hline
       \multirow{3}{*}{2}  &5&\multirow{3}{*}{62.78}&\multirow{3}{*}{53.34}&\multirow{3}{*}{61.73}&\multirow{3}{*}{72.43}  &65.76  &63.96 &64.79 &65.42 &58.40  &73.82  &{\color{blue}\textbf{85.63}} &{\color{red}\textbf{87.85}}    \\
                           & 10         & & & & &65.83  &64.17 &67.43 &65.21 &64.10 &72.99  &{\color{blue}\textbf{89.25}} &{\color{red}\textbf{96.81}}     \\
                           & 25         & & & & &65.76  &67.29 &65.07 &85.63 &64.17 &90.90  &{\color{blue}\textbf{92.78}} &{\color{red}\textbf{94.03}}     \\ \hline
       \multirow{3}{*}{3}  & 5&\multirow{3}{*}{42.29}&\multirow{3}{*}{58.12}&\multirow{3}{*}{51.78}&\multirow{3}{*}{62.63}&39.93  &46.25 &46.04  &39.38 &57.64 &67.71   &{\color{blue}\textbf{73.40}}  &{\color{red}\textbf{89.98}}     \\
                           & 10         & & & & &44.10  &46.32 &46.04 &67.08 &58.75  &79.93  &{\color{blue}\textbf{88.19}} &{\color{red}\textbf{95.63}}     \\
                           & 25         & & & & &42.71  &46.53 &40.97 &85.97 &64.03  &86.04  &{\color{blue}\textbf{95.21}} &{\color{red}\textbf{99.47}}     \\ \hline
    \end{tabular}}
  \end{center}
\end{table*}
To evaluate unsupervised fault diagnosis, three metrics, namely accuracy (ACC), normalized mutual information (NMI), and adjusted rand index (ARI), are used in the experiment. Additionally, all experiments are repeated 10 times, and the three metrics represent the average mean across the 10 repetitions. The proposed method is compared with three related dimension reduction clustering models and three related classification models. These models include the conventional k-means \cite{ref35}, t-SNE \cite{reftsne}, kPCA \cite{refkpca}, LLE \cite{reflle}, PCA \cite{ref16}, AE \cite{refae}, Un-LDA \cite{ref36}, the method proposed by Feng et al. \cite{ref34} using LSVM, NR, and RF. In addition, we established an ablation experiment, the experimental results of the model without graph structure embedding, which we called Our-nGraph.

\begin{figure*}[h]
\center
    \subfigure[Group 1]{\includegraphics[scale=0.345]{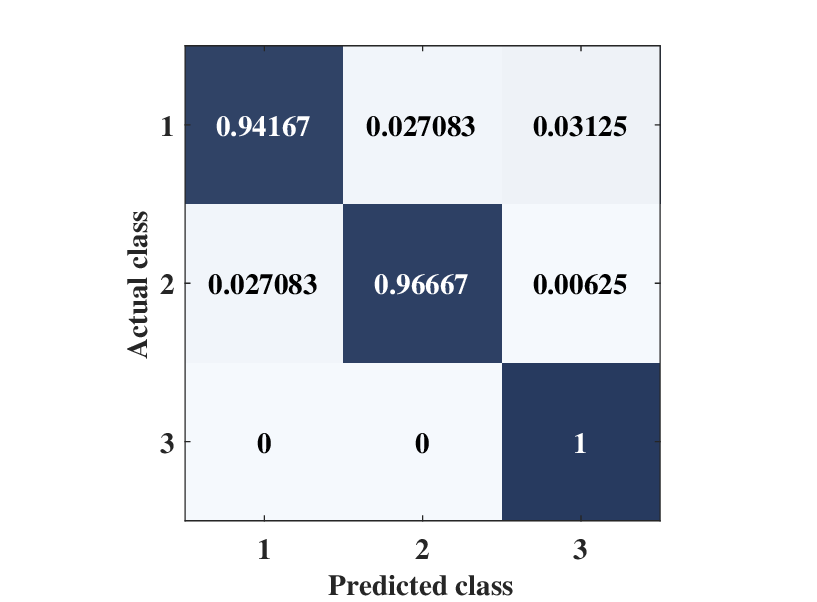}}\hspace{-0.8cm}
    \subfigure[Group 2]{\includegraphics[scale=0.345]{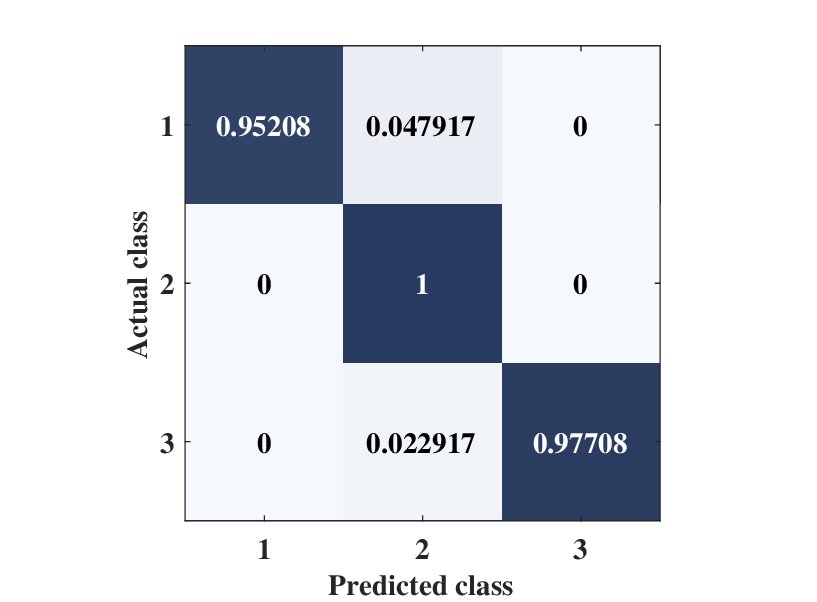}}\hspace{-0.8cm}
    \subfigure[Group 3]{\includegraphics[scale=0.345]{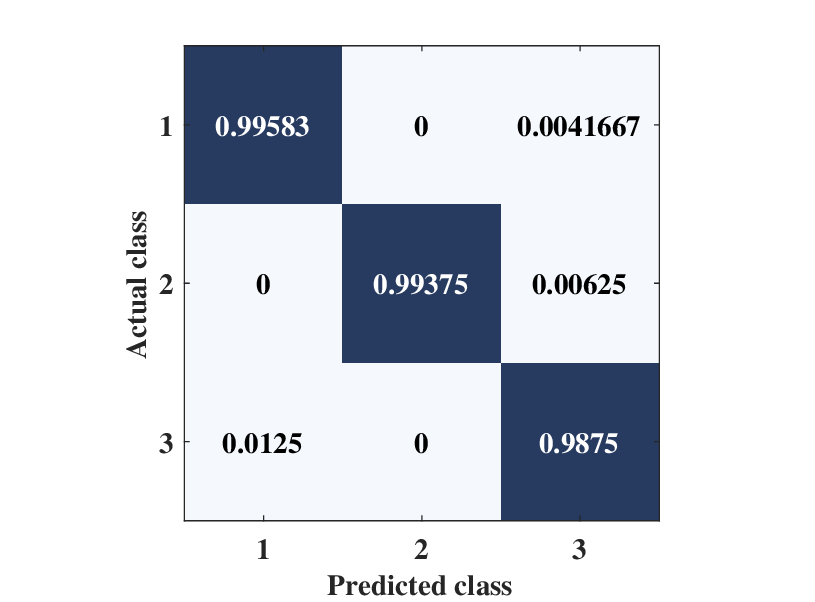}}\hspace{-0.8cm}
    \subfigure[Group 4]{\includegraphics[scale=0.4]{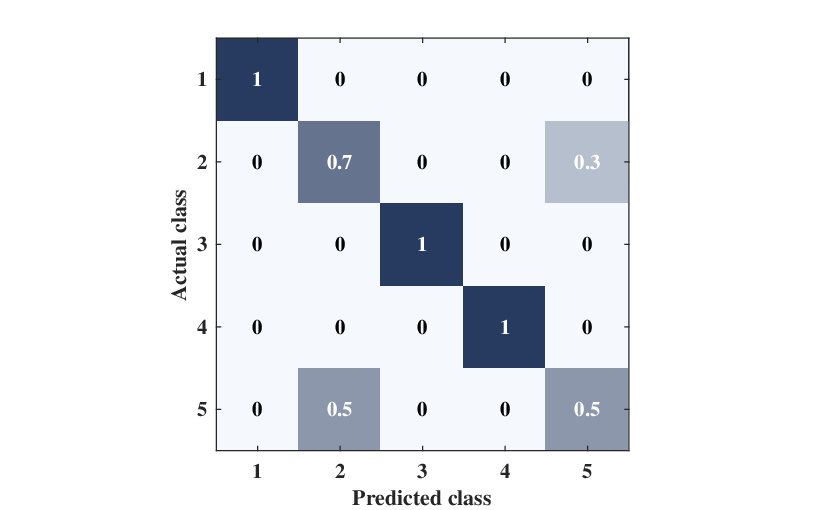}}

    \caption{The confusion matrix of four groups.}\label{HX}
\end{figure*}
\subsection{Performance for High-dimension Data }
To evaluate the dimension reduction performance of the proposed method, we reduced the data from the original dimension to three different smaller dimensions. The reduced dimension is approximately one-tenth of the original dimension.

\begin{table}[h!]
  \begin{center}
    \caption{NMI of the three groups of TE process dataset.}\label{result1nmi}
    \resizebox{\linewidth}{!}{
    \begin{tabular}{ccccccccccc}
    \hline
    \textbf{Groups}    & \textbf{Dim}     &\textbf{K-means}   &\textbf{t-SNE} &\textbf{kPCA} &\textbf{LLE}&\textbf{PCA} &\textbf{AE} &\textbf{Un-LDA}  &\textbf{Ours-nGraph}  &\textbf{Ours}  \\ \hline
       \multirow{3}{*}{1}  & 5         &\multirow{3}{*}{51.23} &{\color{red}\textbf{82.94}}  &50.98 &59.47   &75.08     &47.73   &61.03      &65.40 &{\color{blue}\textbf{78.00}}     \\
                           & 10        &                       &{\color{blue}\textbf{83.79}} &51.13 &63.26   &79.13     &49.21   &57.96      &79.99 &{\color{red}\textbf{88.50}}    \\
                           & 25        &                       &83.25 &52.57 &46.25  &70.14     &74.22   &62.44      &{\color{blue}\textbf{83.95}} &{\color{red}\textbf{90.74}}    \\ \hline
       \multirow{3}{*}{2}  &5          &\multirow{3}{*}{36.47} &37.19 &37.07 &42.52  &46.18     &49.00    &{\color{blue}\textbf{58.95}}      &57.05 &{\color{red}\textbf{67.48}}    \\
                           & 10        &                       &37.20 &37.38 &41.47  &46.02     &42.61    &58.64      &{\color{blue}\textbf{71.08}} &{\color{red}\textbf{87.17}}     \\
                           & 25        &                       &37.19 &40.90 &38.57  &75.29     &46.25    &72.30       &{\color{blue}\textbf{78.43}} &{\color{red}\textbf{81.51}}    \\ \hline
       \multirow{3}{*}{3}  & 5         &\multirow{3}{*}{36.42} &9.24 &12.53  &9.90   &10.22       &57.44  &{\color{blue}\textbf{55.21}}      &42.27 &{\color{red}\textbf{77.37}}     \\
                           & 10         &                      &4.88 &12.68  &10.34  &56.19       &50.34  &66.86      &{\color{blue}\textbf{71.18}} &{\color{red}\textbf{85.85}}     \\
                           & 25         &                      &3.50 &12.78  &6.39   &71.92       &57.03  &71.99      &{\color{blue}\textbf{84.20}} &{\color{red}\textbf{97.35}}     \\ \hline
    \end{tabular}}
  \end{center}
\end{table}

\begin{table}[h!]
  \begin{center}
    \caption{PUR of the three groups of TE process dataset.}\label{result1pur}
    \resizebox{\linewidth}{!}{
    \begin{tabular}{ccccccccccc}
    \hline
    \textbf{Groups}    & \textbf{Dim}     &\textbf{K-means}    &\textbf{t-SNE} &\textbf{kPCA}&\textbf{LLE} &\textbf{PCA}  &\textbf{AE} &\textbf{Un-LDA}  &\textbf{Ours-nGraph}  &\textbf{Ours}  \\ \hline
       \multirow{3}{*}{1}  & 5         &\multirow{3}{*}{71.81} &{\color{red}\textbf{94.51}}  &77.92 &83.54  &92.01      &71.11  &76.81      &87.15 &{\color{blue}\textbf{93.96}}     \\
                           & 10        &                       &{\color{blue}\textbf{94.86}} &78.06 &87.43  &93.98      &65.49  &67.57      &92.47 &{\color{red}\textbf{97.22}}     \\
                           & 25        &                       &94.58 &77.99 &71.81  &89.61      &90.28  &84.72      &{\color{blue}\textbf{96.78}} &{\color{red}\textbf{97.85}}     \\ \hline
       \multirow{3}{*}{2}  & 5         &\multirow{3}{*}{62.78} &65.76 &63.96 &64.79  &65.42     &64.44   &73.82      &{\color{blue}\textbf{85.63}} &{\color{red}\textbf{87.85}}    \\
                           & 10        &                       &65.83 &64.17 &67.43  &65.21     &64.10   &72.99      &{\color{blue}\textbf{89.25}} &{\color{red}\textbf{96.81}}     \\
                           & 25        &                       &65.76 &67.29 &65.07  &85.63     &64.44   &90.90      &{\color{blue}\textbf{92.78}} &{\color{red}\textbf{94.03}}     \\ \hline
       \multirow{3}{*}{3}  & 5         &\multirow{3}{*}{42.29} &44.72 &45.49 &46.25  &39.38     &66.60   &67.71      &{\color{blue}\textbf{73.61}} &{\color{red}\textbf{89.98}}     \\
                           & 10         &                      &44.10 &46.32 &46.04  &67.08     &64.86   &79.93      &{\color{blue}\textbf{88.19}} &{\color{red}\textbf{95.63}}     \\
                           & 25         &                      &42.71 &46.53 &40.97  &85.97     &66.53   &86.04      &{\color{blue}\textbf{95.21}} &{\color{red}\textbf{99.47}}     \\ \hline
    \end{tabular}}
  \end{center}
\end{table}

For the TE process dataset, the dimension is reduced to 5, 10, and 25 from the original dimension of 52. Table \ref{result1acc} presents the fault diagnosis accuracy of the proposed method after dimension reduction, as well as the fault diagnosis accuracy of the methods compared. The best results are highlighted in bold and marked with red color, while the second-best result is highlighted in bold and marked with blue. Tables \ref{result1acc}, \ref{result1nmi}, and \ref{result1pur} the ACC, NMI, and PUR scores of the proposed method and the clustering methods. It is evident that as the dimension reduction increases, the diagnostic accuracy tends to improve. Our method achieved the highest accuracy across three groups, except for the first group where, when reduced to 5 dimensions, the t-SNE method obtained the highest accuracy, NMI, and PUR. Additionally, our proposed method achieved the second-highest results in the second and third groups when graph structure embedding were not used.

In the case of the HSM dataset, the dimension is reduced to 30, 50, and 100 as the original dimension is 1000. The three diagnostic indicators are presented in Table \ref{result2}, where they are also compared with the aforementioned clustering methods, excluding t-SNE. This exclusion is due to memory constraints encountered when reducing dimensions to 30, 50, and 100. Our method achieves the best results on all three metrics, while Un-LDA achieves the second-best results. It is noteworthy that our method achieves approximately 10\% higher accuracy (ACC) compared to Un-LDA, and also higher NMI (0.1\%) and PUR (3.93\%) scores than their method. In comparison, the other methods achieve only half of our performance in these metrics.

\begin{table}[h!]
  \begin{center}
    \caption{Performance of HSM dataset.}\label{result2}
        \resizebox{\linewidth}{!}{
    \begin{tabular}{ccccccccccc}
    \hline
  \textbf{Metrics}     &\textbf{Dim}  &\textbf{K-means}       &\textbf{kPCA} &\textbf{LLE} &\textbf{PCA} &\textbf{AE} &\textbf{Un-LDA} &\textbf{Ours-nGraph}   &\textbf{Ours}  \\ \hline
    \multirow{3}{*}{ACC} & 30         &\multirow{3}{*}{34.00} &72.00 &36.00    &42.00        &50.00     &{\color{blue}\textbf{74.00}}          & 60.00    &{\color{red}\textbf{83.20}}     \\
                         & 50         &                       &72.00 &36.00    &27.20        &56.00     &{\color{blue}\textbf{73.00}}          & 65.00    &{\color{red}\textbf{82.16}}     \\
                         & 100        &                       & -    & -       &25.20        &62.00     &72.00          &{\color{blue}\textbf{ 74.00}}    &{\color{red}\textbf{86.00}}    \\ \hline
    \multirow{3}{*}{NMI} & 30         &\multirow{3}{*}{22.44} &78.61 &10.17    &46.34        &54.48     &{\color{blue}\textbf{82.77}}          & 65.55    &{\color{red}\textbf{83.44}}    \\
                         & 50         &                       &78.61 &19.43    &13.18        &65.55     &{\color{blue}\textbf{82.77}}          & 63.78    &{\color{red}\textbf{82.99}}    \\
                         & 100        &                       & -    & -       &10.00        &61.96     &{\color{blue}\textbf{82.77}}          & 74.60    &{\color{red}\textbf{84.35}}     \\ \hline
   \multirow{3}{*}{PUR}  & 30         &\multirow{3}{*}{38.00} &78.00 &34.00    &46.60        &54.00     &{\color{blue}\textbf{80.00}}          & 60.00    &{\color{red}\textbf{83.60}}     \\
                         & 50         &                       &78.00 &36.00    &31.60        &60.00     &{\color{blue}\textbf{80.00}}          & 67.80    &{\color{red}\textbf{82.20}}     \\
                         & 100        &                       & -    & -       &29.40        &64.00     &{\color{blue}\textbf{80.00}}          & 75.20    &{\color{red}\textbf{86.00}}     \\ \hline
    \end{tabular}}
  \end{center}
\end{table}

To better show the performance of our method, the confusion matrix of the four-groups experiments is shown in Fig. \ref{HX}. From the confusion matrix, it becomes evident that the diagnosis accuracy of the first three groups is very close to 100\% not only the average accuracy but also the diagnosis accuracy of each category. In the case of the HSM dataset, the fault diagnosis accuracy for fault types 1, 3, and 4 is 100\%, indicating accurate identification of these faults. However, for fault types 2 and 5, the accuracy is 70\% and 50\% respectively, indicating some misclassification. The errors in fault diagnosis are mainly observed for fault types 2 and 5. Specifically, 30\% of fault type 2 instances are incorrectly classified as fault type 5, while 50\% of fault type 5 instances are mistakenly classified as fault type 2. This misclassification can be attributed to the similarities in the extracted features between these two fault types, which may be due to the limited amount of available data.

\subsection{Performance of Outliers or Noisy Data}
Noise is a typical and widespread disturbance in the industrial process. This section seeks to analyze their impact on the performance of our method and evaluate its robustness in the presence of noise. Specifically, we explore two types of noise: outliers and signal noise. The experiment regarding outliers is conducted using the TE process dataset, while the experiment involving signal noise is carried out using the HSM dataset.

The value for defining outliers follows the experiments described by Xie et al. \cite{ref37}, which is 1.5 times the value of the normal data. Table \ref{Outlier} shows the diagnosis accuracy of TE process data under three different outliers proportions: outlier-free, 1\% outliers, and 5\% outliers. From Table \ref{Outlier}, it is clear that as the proportion of outliers increases, the diagnostic accuracy of all four methods decreases. However, our method consistently outperforms the other methods and achieves the highest diagnostic accuracy across all three groups of experiments. Furthermore, when the proportion of outliers is 5\%, the diagnostic accuracy of the other methods ranges from 33.47\% to 73.89\%. In contrast, our method achieves higher diagnostic accuracy ranging from 63.32\% to 79.31\% across the three groups. This demonstrates that our method exhibits better robustness in the presence of outliers compared to the other methods.
\begin{table}[h!]
  \begin{center}
    \caption{Diagnosis accuracy of the proposed method under different proportions of outliers in TE process dataset}\label{Outlier}
      \resizebox{\linewidth}{!}{
    \begin{tabular}{ccccccccccc}
    \hline
  \textbf{Groups}     &\textbf{Outliers}  &\textbf{K-means}  &\textbf{t-SNE} &\textbf{kPCA}  &\textbf{LLE}     &\textbf{PCA}  &\textbf{AE} &\textbf{Un-LDA}  &\textbf{Ours-nGraph}  &\textbf{Ours}  \\ \hline
    \multirow{3}{*}{1}   & 0              &71.81             &94.58 &77.99  &71.81     &92.98     &90.28  &84.72  &{\color{blue}\textbf{96.78}}    &{\color{red}\textbf{97.85}}     \\
                         & 1\%            &42.29             &82.17 &77.15  &61.11     &{\color{blue}\textbf{74.58}}     &60.35  &72.57  &73.84    &{\color{red}\textbf{82.08}}     \\
                         & 5\%            &40.21             &70.07 &{\color{blue}\textbf{73.89}}  &60.49     &66.72     &60.07  &41.67  &72.15    &{\color{red}\textbf{78.65}}    \\ \hline
    \multirow{3}{*}{2}   & 0              &62.78             &65.76 &67.36  &64.07     &85.63     &64.17  &90.90  &{\color{blue}\textbf{92.78}}    &{\color{red}\textbf{94.03}}    \\
                         & 1\%            &58.33             &65.49 &67.50  &62.22     &60.56     &60.21  &{\color{blue}\textbf{74.58}}  &71.27    &{\color{red}\textbf{79.31}}    \\
                         & 5\%            &35.35             &59.65 &66.74  &60.90     &33.47     &59.31  &34.51  &{\color{blue}\textbf{64.10}}    &{\color{red}\textbf{66.74}}     \\ \hline
   \multirow{3}{*}{3}    & 0              &42.29             &42.71 &46.53  &40.97     &85.97     &64.03  &86.04  &{\color{blue}\textbf{95.21}}    &{\color{red}\textbf{99.47}}     \\
                         & 1\%            &40.28             &38.26 &45.97  &38.06     &74.86     &39.79 &70.42   &{\color{blue}\textbf{86.37}}    &{\color{red}\textbf{92.15}}     \\
                         & 5\%            &36.18             &47.57 &45.28  &34.38     &33.40    &43.61 &38.54    &{\color{blue}\textbf{47.99}}    &{\color{red}\textbf{63.32}}     \\ \hline
    \end{tabular}}
  \end{center}
\end{table}

Since the HSM data is collected from the current signal of the motor and reflects the state of the roller, which can be considered as bearing data, it is reasonable to expect the presence of signal noise in the collected data. To evaluate the performance under different signal-to-noise ratios (SNR), we set the SNR to 0 dB, 5 dB, 10 dB, 15 dB, and 20 dB. Table \ref{Signal} displays the diagnostic accuracy of our method and the other methods under these different SNR values. From the table, it is evident that as the SNR increases, the diagnostic accuracy decreases. However, the other methods are unable to effectively handle the data with signal noise, as their diagnostic accuracy remains similar across different SNR levels. In contrast, our method demonstrates higher diagnostic accuracy at low SNR compared to high SNR. This proves the effectiveness of our method in diagnosing faults under signal noise.

\begin{table}[h!]
  \begin{center}
    \caption{Diagnosis accuracy of the proposed method under different signal noise ratio in HSM dataset}\label{Signal}
          \resizebox{\linewidth}{!}{
    \begin{tabular}{ccccccccc}
    \hline
                         \textbf{Noisy}    &\textbf{K-means}   &\textbf{kPCA}  &\textbf{LLE}     &\textbf{PCA}  &\textbf{AE}&\textbf{Un-LDA}  &\textbf{Ours-nGraph}  &\textbf{Ours}  \\  \hline
                          0dB              &34.00                           &72.00          &36.00            &42.00         &62.00      &{\color{blue}\textbf{74.00}}  &{\color{blue}\textbf{74.00}}  &{\color{red}\textbf{86.00}}     \\
                          5dB              &33.00                           &58.00          &34.20            &38.90         &42.00      &42.00  &{\color{blue}\textbf{65.40}}  &{\color{red}\textbf{75.42}}     \\
                          10dB             &30.40                           &{\color{blue}\textbf{54.80}}          &34.60            &37.90         &36.00      &42.00  &44.00  &{\color{red}\textbf{63.66}}    \\
                          15dB             &28.20                           &{\color{blue}\textbf{51.20}}          &33.60            &36.30         &46.00      &42.00  &42.00  &{\color{red}\textbf{55.84}}    \\
                          20dB             &29.80                           &{\color{blue}\textbf{48.60}}          &32.20            &36.40         &42.00      &42.00  &42.00  &{\color{red}\textbf{50.71}}    \\

            \hline
    \end{tabular}}
  \end{center}
\end{table}

\subsection{Parameter Sensitivity and Convergence}
\begin{figure*}[h]
\center
    \subfigure[Group 1]{\includegraphics[scale=0.3]{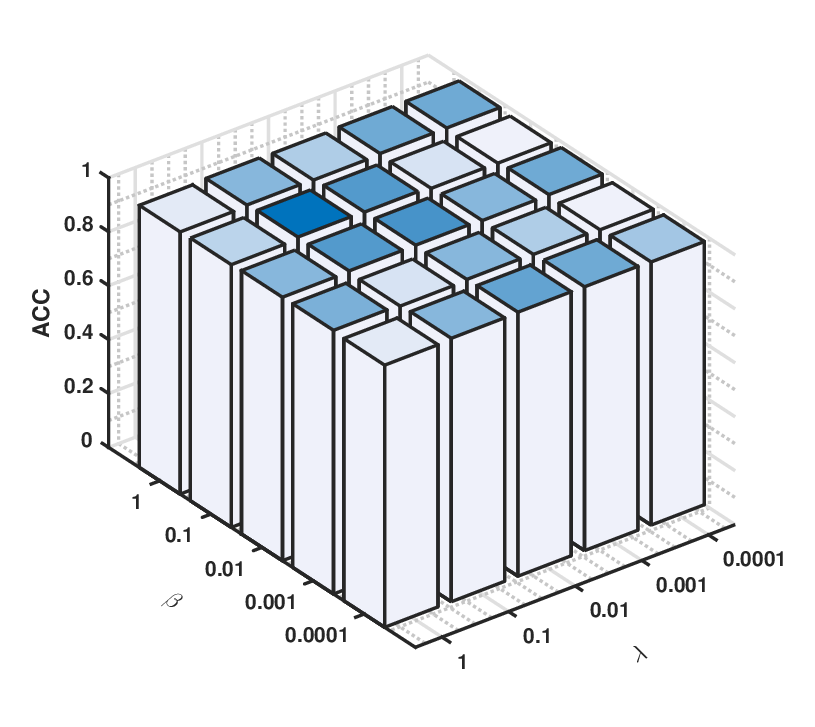}}\hspace{-0.4cm}
    \subfigure[Group 2]{\includegraphics[scale=0.3]{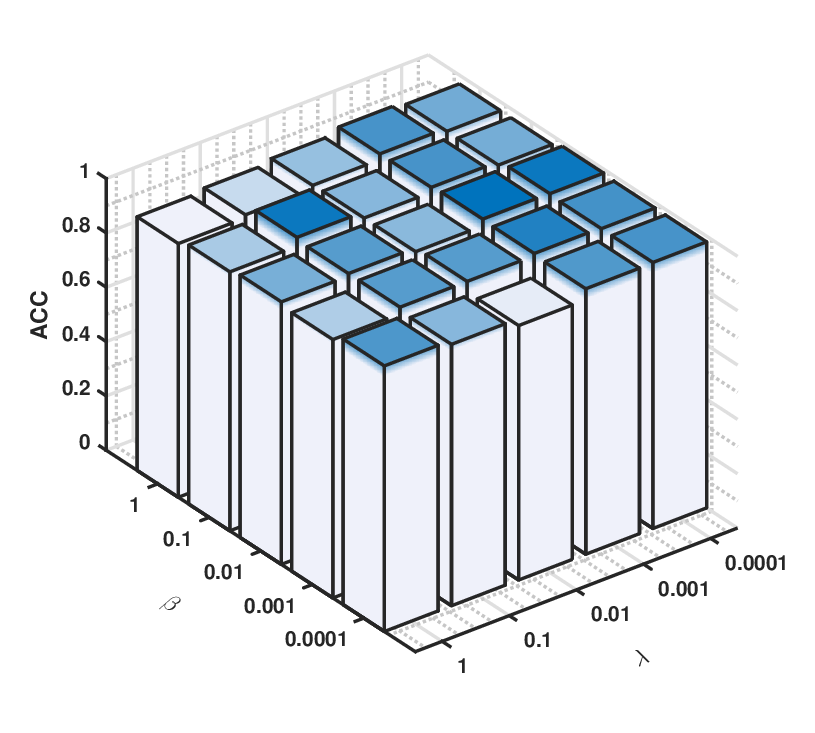}}\hspace{-0.4cm}
    \subfigure[Group 3]{\includegraphics[scale=0.3]{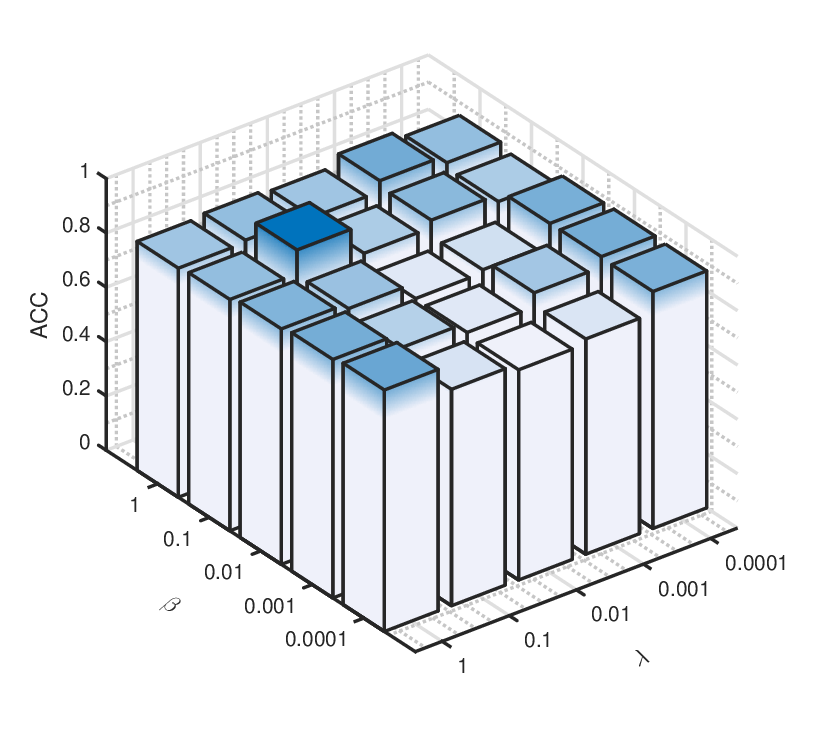}}\hspace{-0.4cm}
    \subfigure[Group 4]{\includegraphics[scale=0.3]{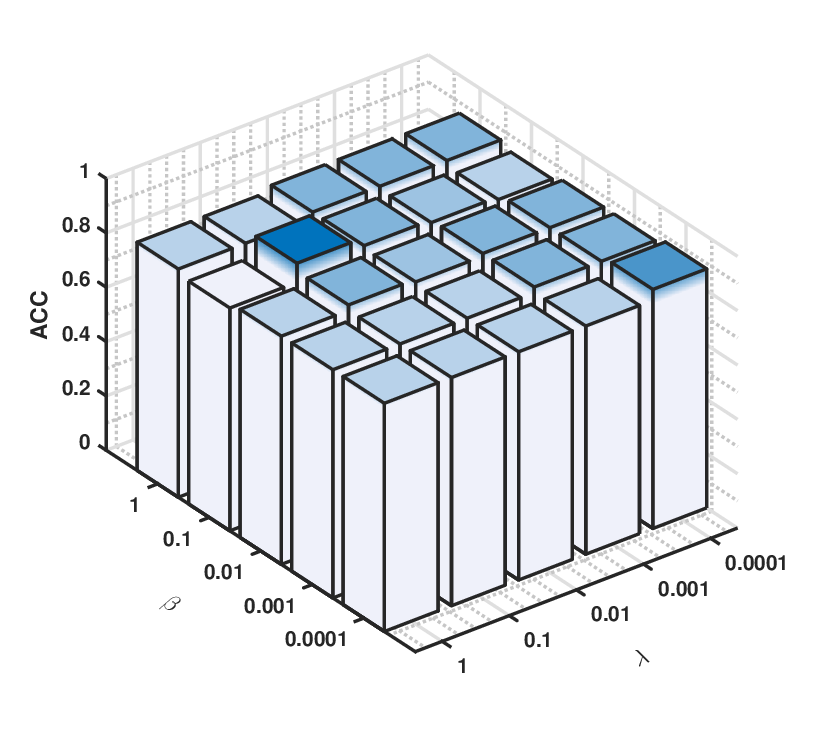}}\hspace{-0.4cm}
    \caption{Diagnosis accuracy of four groups for different hyperparameters.}\label{Para}
\end{figure*}

In this section, we examine the impact of hyperparameters $\lambda$ and $\beta$ on the performance of fault diagnosis, while also demonstrating the convergence of the proposed optimal algorithm. The values of hyperparameters $\lambda$ and $\beta$ are set to $[1, 0.1, 0.01, 0.001, 0.0001]$. Fig. \ref{Para} shows the diagnostic accuracy of the four groups under different parameter values. The color of the graph represents the diagnostic accuracy, with darker shades indicating higher accuracy. From the figure, we observe that the diagnostic accuracy fluctuates within a small range across different parameter values. The best results are obtained when $\lambda = 0.1$ and $\beta = 0.1$, particularly for groups 3 and 4.

\begin{figure*}[h]
\center
    \subfigure[F]{\includegraphics[scale=0.3]{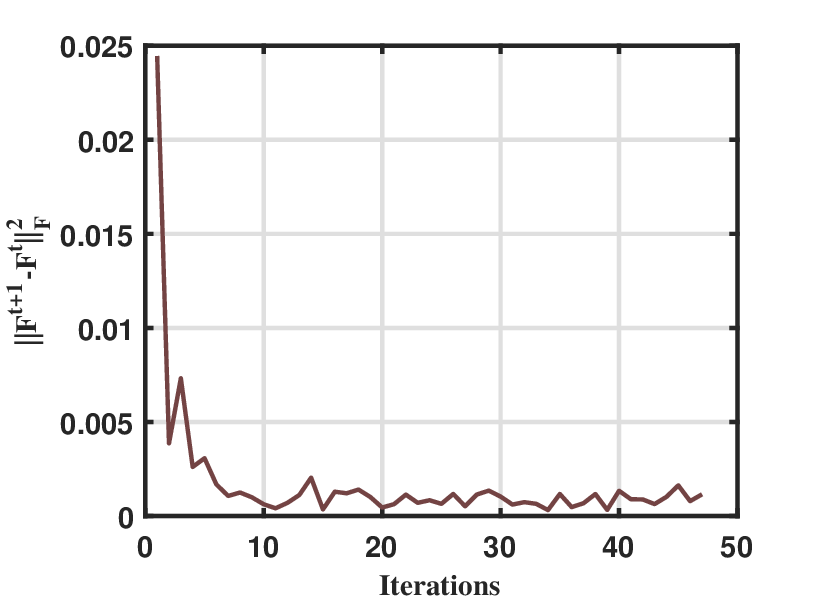}}\hspace{-0.4cm}
    \subfigure[W]{\includegraphics[scale=0.3]{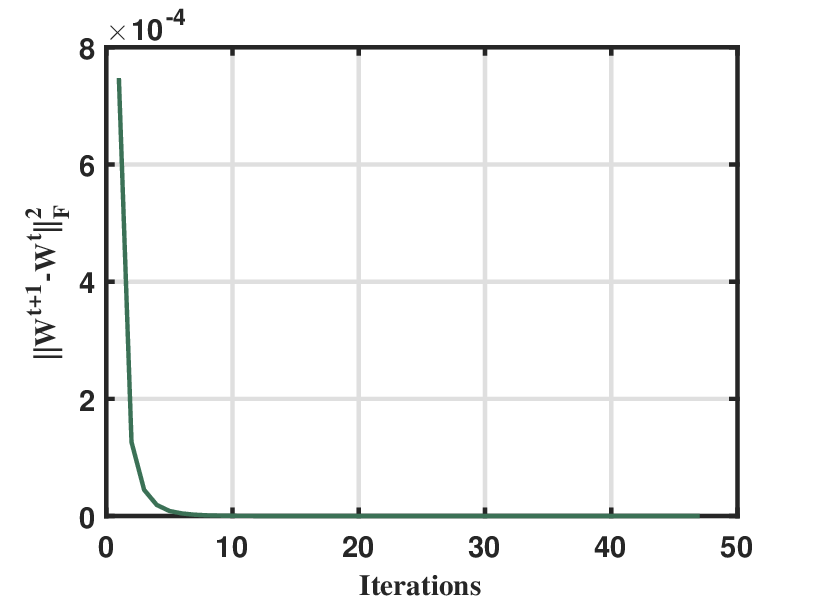}}\hspace{-0.4cm}
    \subfigure[S]{\includegraphics[scale=0.3]{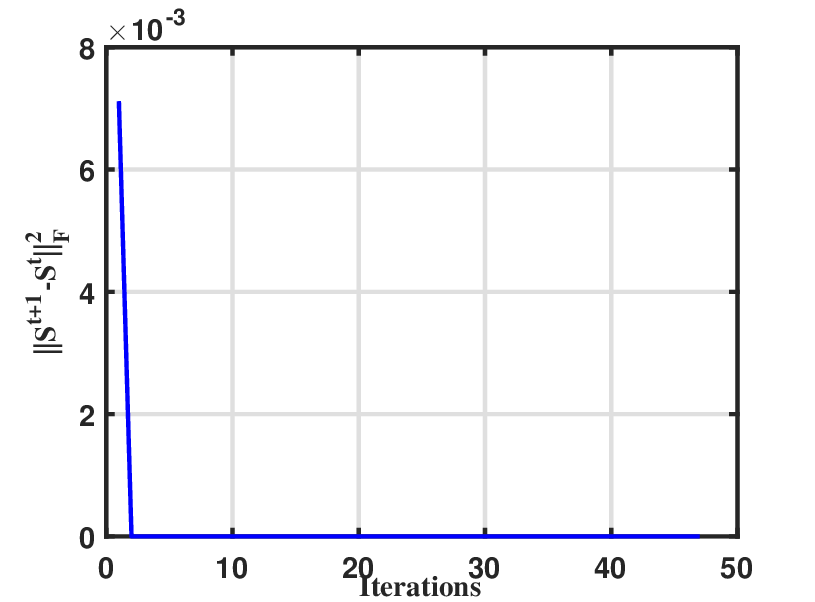}}\hspace{-0.4cm}
    \subfigure[G]{\includegraphics[scale=0.3]{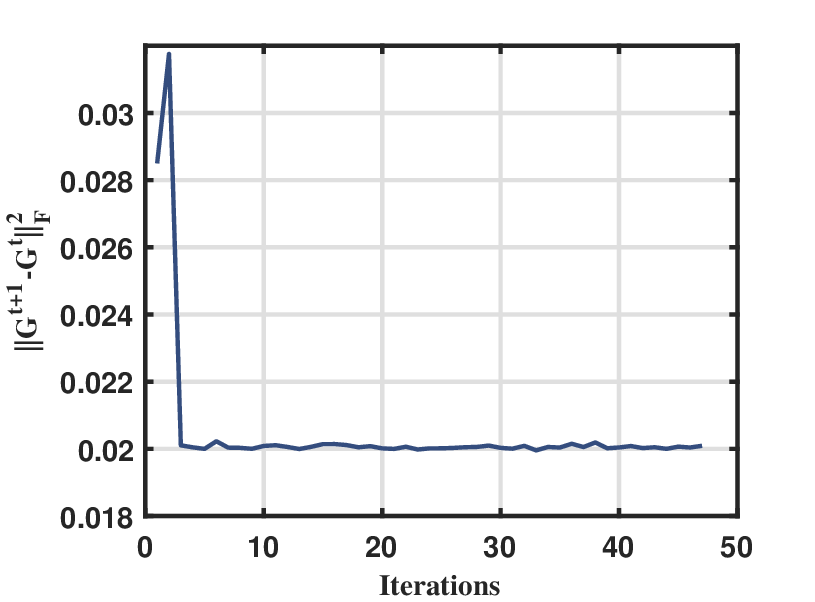}}\hspace{-0.4cm}
    \caption{The convergence of the four variables.}\label{Conv}
\end{figure*}

Additionally, Fig. \ref{Conv} shows the convergence of the four variables that require optimization. It is evident that as the number of iterations increases, the variables converge rapidly. In fact, all the variables reach convergence within 10 iterations. These experimental results align with the theoretical analysis presented in Section 3.2, providing further evidence that our optimization algorithm converges to the desired solution.

\section{Conclusion}
In this paper, a new robust unsupervised fault diagnosis method is proposed for high-dimensional and noisy data. The proposed method reduces the dimension using projected fault data onto a dimension-reduced space and extracts fault features from the graph structure. Furthermore, the proposed method handles noise from the respective model optimization, and the $l_{2,1}$-norm is introduced into the objective model for dropping the influence of noise, as well as the typicality-aware constraint. Experiments on the benchmark TE process and the real hot roller of steel process show the effectiveness of the proposed method, which can deal with at least two types of noise when reducing the dimension of fault data and achieve the best results compared with the existing fault clustering methods.


\begin{IEEEbiography}[{\includegraphics[width=1in,height=1.2in,clip,keepaspectratio]{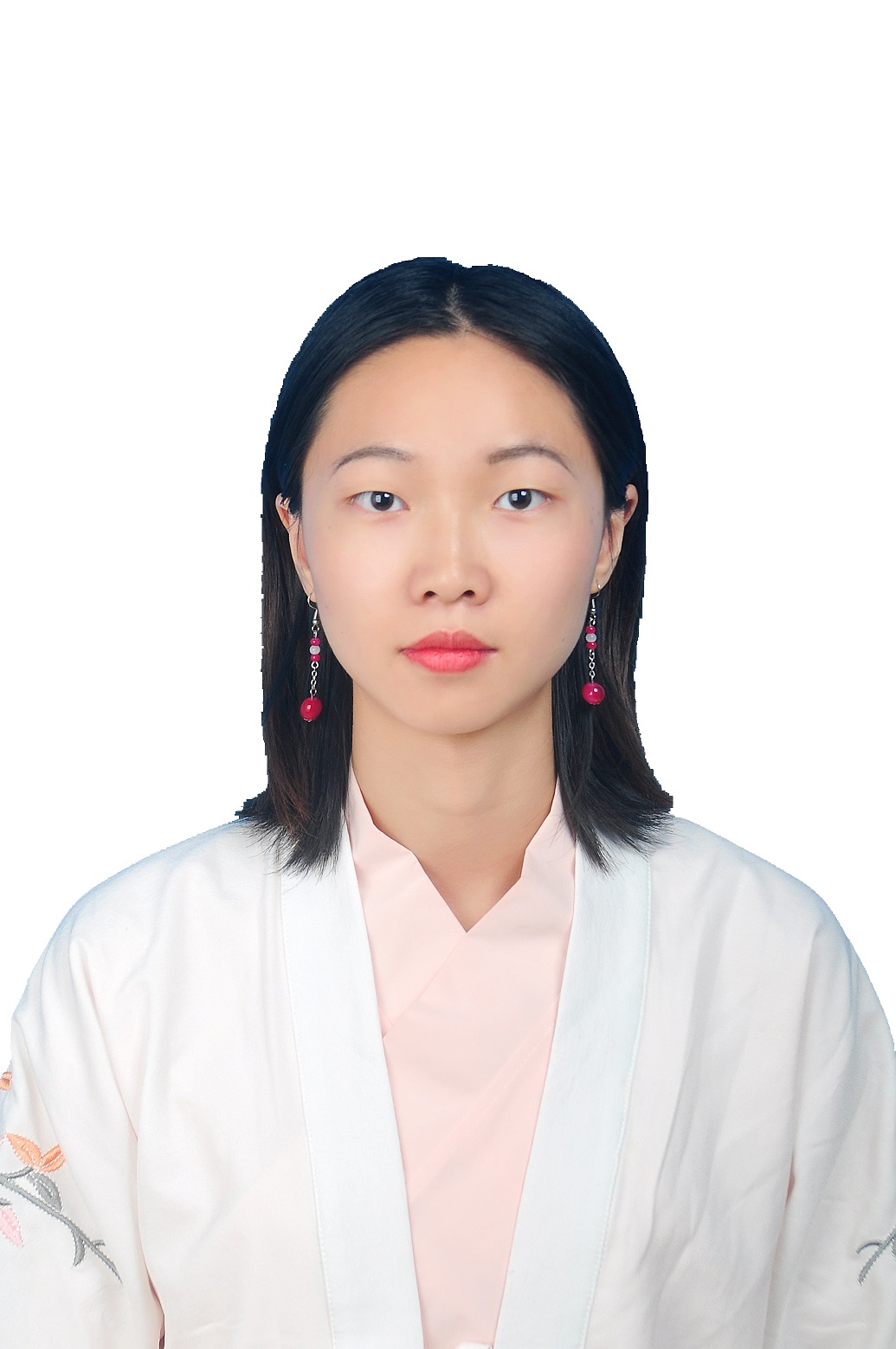}}]{Dandan Zhao}
received her B.E. degree at the Faculty of Information Engineering and Automation from Kunming University of Science and Technology, China, in 2019. She is currently working towards a doctorate in artificial intelligence from the College of Automation, Chongqing University.

Her research interests include fault diagnosis, deep learning, and machine learning.
\end{IEEEbiography}

\vspace{-11pt}
\begin{IEEEbiography}[{\includegraphics[width=1in,height=1.2in,clip,keepaspectratio]{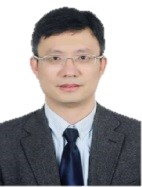}}]{Hongpeng Yin}
received his B.E. degree and Ph.D. degree at College of Automation, Chongqing University, Chongqing, China, in 2003 and 2009, respectively.

He is currently a professor at the College of Automation, Chongqing University, Chongqing, China. His research interests include advanced data-driven approaches for diagnostics and prognostics, including machine learning, signal processing, and information fusion.
\end{IEEEbiography}

\vspace{-11pt}

\begin{IEEEbiography}[{\includegraphics[width=1in,height=1.2in,clip,keepaspectratio]{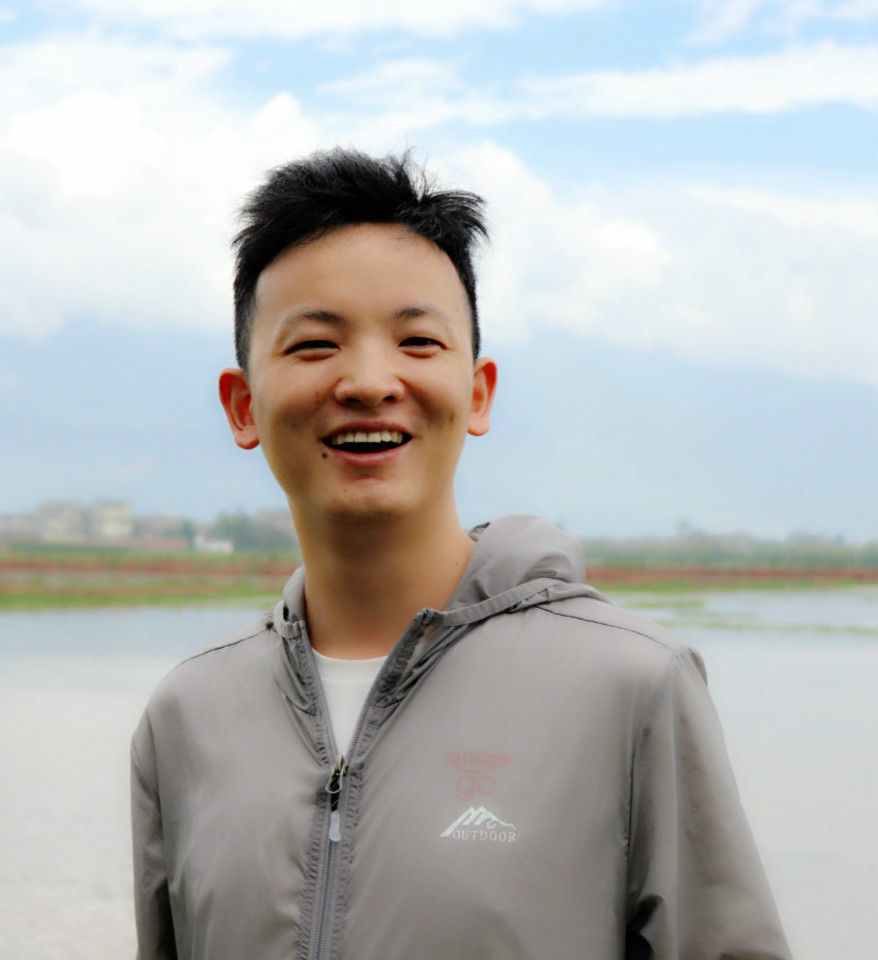}}]{Jingtang Bian}
received hid master’s degree in computer science from the Northwestern Polytechnical University in 2021. He is currently pursuing hid doctorate at Sun Yat-sen University.

His research interests are machine learning and pattern recognition. His publications appear in IEEE Transactions on Pattern Analysis and Machine Intelligence, IEEE Transactions on Neural Networks and Learning Systems, IEEE Transactions on Knowledge and Data Engineering, etc.
\end{IEEEbiography}

\vspace{-11pt}
\begin{IEEEbiography}[{\includegraphics[width=1in,height=1.2in,clip,keepaspectratio]{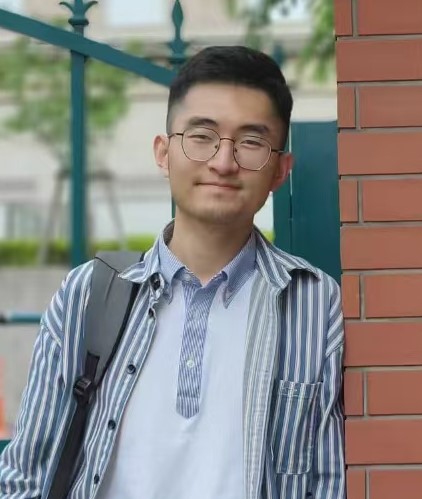}}]{Han Zhou}
is currently a research scientist at Peng Cheng Laboratory, Shenzhen, China. He received his Ph.D. degree from Chongqing University, Chongqing, China, in 2024. During his PhD, he was a special research student at the University of Tokyo, Tokyo, Japan, from 2021 to 2023.

His research interests include XX and online learning.
\end{IEEEbiography}

\vfill

\end{document}